\documentclass[nohyperref]{article}

\usepackage{microtype}
\usepackage{graphicx}
\usepackage{subfigure}
\usepackage{booktabs} 

\usepackage[accepted]{icml2022}

\usepackage{amsmath}
\usepackage{amssymb}
\usepackage{mathtools}
\usepackage{amsthm}

\usepackage{graphicx}
\usepackage{graphics}
\usepackage{booktabs}
\usepackage{amsmath}
\usepackage{amssymb}
\usepackage{hyperref}
\usepackage{xcolor}
\usepackage{amsmath}
\usepackage{array}
\usepackage{amssymb}
\usepackage{eucal}

\usepackage{subfigure}
\usepackage{caption}

\usepackage{multirow}
\usepackage{tabularx}
\usepackage{hyperref}

\usepackage[capitalize,noabbrev]{cleveref}

\theoremstyle{plain}

\theoremstyle{definition}

\theoremstyle{remark}

\usepackage[textsize=tiny]{todonotes}

\icmltitlerunning{Continuous-Time Modeling of Counterfactual Outcomes Using Neural Controlled Differential Equations}

\begin{document}

\twocolumn[
\icmltitle{Continuous-Time Modeling of Counterfactual Outcomes Using Neural Controlled Differential Equations}



\icmlsetsymbol{equal}{*}
\begin{icmlauthorlist}
\icmlauthor{Nabeel Seedat}{equal,cam}
\icmlauthor{Fergus Imrie}{equal,ucla}
\icmlauthor{Alexis Bellot}{columbia}
\icmlauthor{Zhaozhi Qian}{cam}
\icmlauthor{Mihaela van der Schaar}{cam,ucla,at}
\end{icmlauthorlist}

\icmlaffiliation{cam}{Department of Applied Mathematics and Theoretical Physics, University of Cambridge, UK}
\icmlaffiliation{at}{The Alan Turing Institute, London, UK}
\icmlaffiliation{ucla}{University of California, Los Angeles, USA}
\icmlaffiliation{columbia}{Columbia University, USA}

\icmlcorrespondingauthor{Nabeel Seedat}{ns741@cam.ac.uk}
\icmlkeywords{Machine Learning, ICML, Counterfactual Estimation, Treatment Effects, Neural Differential Equations, Time Series}

\vskip 0.3in
]



\printAffiliationsAndNotice{\icmlEqualContribution} 

\begin{abstract}

Estimating counterfactual outcomes over time has the potential to unlock personalized healthcare by assisting decision-makers to answer ``what-if” questions. Existing causal inference approaches typically consider regular, discrete-time intervals between observations and treatment decisions and hence are unable to naturally model irregularly sampled data, which is the common setting in practice. To handle arbitrary observation patterns, we interpret the data as samples from an underlying continuous-time process and propose to model its latent trajectory explicitly using the mathematics of controlled differential equations. This leads to a new approach, the Treatment Effect Neural Controlled Differential Equation (TE-CDE), that allows the potential outcomes to be evaluated at any time point. In addition, adversarial training is used to adjust for time-dependent confounding which is critical in longitudinal settings and is an added challenge not encountered in conventional time-series. To assess solutions to this problem, we propose a controllable simulation environment based on a model of tumor growth for a range of scenarios with irregular sampling reflective of a variety of clinical scenarios. TE-CDE consistently outperforms existing approaches in all simulated scenarios with irregular sampling.
\end{abstract}

\section{Introduction}\label{sec:intro}
Decision-makers must answer several critical questions before taking an action. In the clinical setting, before a treatment is given, clinicians must evaluate whether a treatment should be given and, if so, both \textit{what} treatment is best for their patient and \textit{when} the treatment should be administered. Answering such questions requires reliably estimating the effect of a treatment or sequence of treatments. While from a causal inference perspective, clinical trials represent the gold standard to answer these questions, it is highly desirable to estimate treatment effects from observational data. This is due to the significant expense, relatively small sample sizes, and narrow inclusion criteria of clinical trials.

There are several causal inference methods proposed in the static setting (e.g. \citealp{shalit2017estimating,alaa2017bayesian,yoon2018ganite}). However, estimating the effects of treatments \textit{over time} is of paramount importance for real-world administration of complex treatment plans and personalized healthcare.  Only in the longitudinal setting can we understand how diseases evolve under different treatment plans, how individual patients respond to treatment over time, or the optimal timing for treatment.

\begin{figure*}[t]
    \centering
    \includegraphics[width=0.95\textwidth]{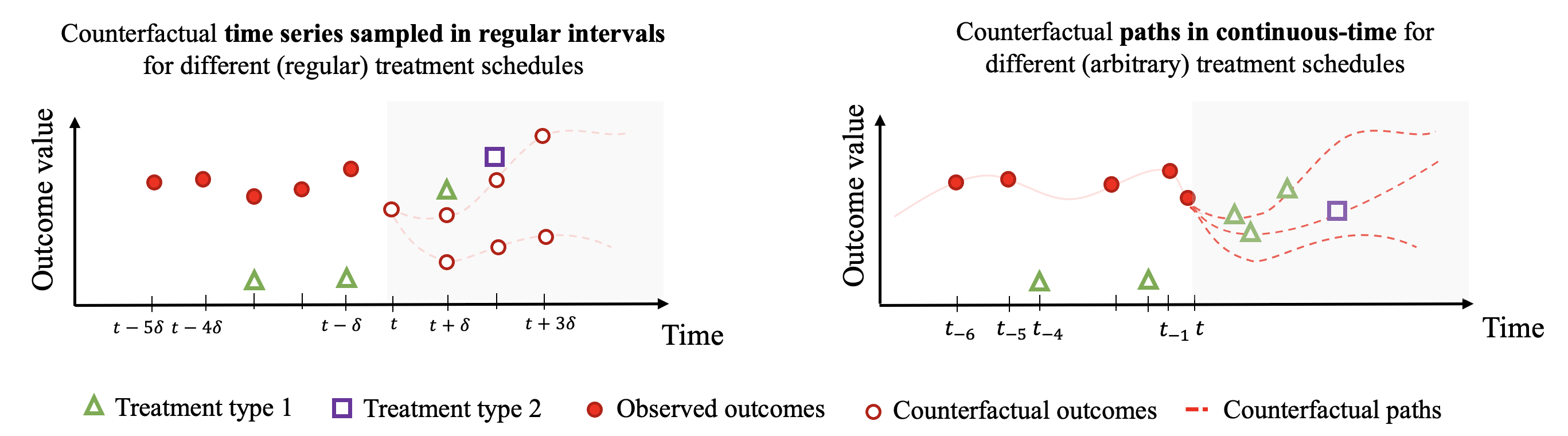}
    \caption{Illustration of the different paradigms of longitudinal data processing. We contrast the regular sampled setting (\textit{left}) which RNN-based methods assume vs the irregularly sampled setting (\textit{right}) which TE-CDE addresses, where data can be observed and evaluations carried out at any time-step.}
    \label{fig:trajectories}
\end{figure*}

However, estimating counterfactual outcomes in the longitudinal setting introduces additional challenges, the most significant of which is that the observed treatment assignment may depend on confounding variables that vary over time (time-dependent confounding, \citealp{platt2009time}). For example, not all cancer patients are equally likely to be offered the same chemotherapy regimen. 
In particular, the history of patients' covariates and their response to past treatments affects future treatments \cite{bica2021real}. This can introduce bias in causal effects and variance in the estimation of counterfactuals due to the systematic differences in the distribution of confounding variables between any two sets of treatments over time. 

This issue of time-dependent confounding and distribution shift is the primary challenge of causal inference over time, not encountered in standard time-series. Hence, conventional time-series models are not applicable to our setting as they do not adjust for bias introduced by time-varying confounders and hence are sensitive to the policy in the observational data \cite{schulam2017reliable}. 

While prior work in causal inference has sought to mitigate such confounding bias \cite{robins2000marginal,lim2018forecasting,bica2020estimating}, the setting considered is overly restrictive and does not reflect most real-world observation data. In particular, previous work assumes that data is regular and arrives at fixed, evenly spaced time intervals and that the sampling times perfectly coincide between different individuals. However, neither is true in practice, significantly limiting the practical use of such methods. 

Discretizing the patient's evolution over time, an inherently continuous process, has significant limitations, both when learning from historical data and for prospective clinical use.
From a learning perspective, observational data is typically not sampled regularly.
Indeed, irregularity in observational data can manifest for simple reasons, such as scheduling, a patient missing an appointment, or a healthcare practitioner not capturing the observation, to more complex considerations, for example more severe cases are often observed more frequently while different treatments can require differences in monitoring.

Prospective use cases raise similar issues surrounding mismatches between the discretization scheme and desired evaluation times that means the chosen discretization may not be applicable.
As a result, for real-world applications where data is sampled irregularly, we believe that treatment effects over time should be modeled in a continuous manner.

\textbf{Contributions.} In this paper, we address the realistic but understudied problem of counterfactual estimation in the irregularly sampled setting with time-dependent confounding; a significantly more complex setting for counterfactual estimation than the standard regular, discrete setting.

To do so, we depart from existing methods based on recurrent neural networks (RNNs) and propose a novel alternative inspired by recent breakthroughs in neural controlled differential equations (CDEs) \cite{kidger2020neural}, which we call the \textit{Treatment Effect Neural Controlled Differential Equation (TE-CDE)}.

To model the observation histories, we learn a \textit{continuous latent representation} of the patient state as the solution to a CDE. To the best of our knowledge, this is the first work to frame the evolution of a patient's latent state as the solution to a CDE.
This framing enables TE-CDE to learn from arbitrary historical observation patterns and allows potential outcomes to be evaluated at any point in time. 

In addition, we introduce a controllable simulation environment based on a realistic model for tumor growth to generate irregularly sampled observational data.
We demonstrate that the unrealistic assumptions imposed by existing state-of-the-art models lead to reduced performance in a range of irregularly sampled scenarios, and that TE-CDE outperforms these methods across all scenarios with irregularly sampled observation histories.

\section{Related Work}\label{sec:related}

This paper primarily engages with the literature on treatment effect estimation with time-varying covariates, treatments, and outcomes,
but also draws on insights from causality in dynamical systems and recent work on modeling controlled differential equations. We explicitly note the difference between causal inference over time and conventional time series modeling as outlined in Section 1 and hence do not focus on recent advances in time series models. An extended discussion of related work can be found in Appendix \ref{appendixA}. In Table \ref{contrast}, we contrast the problem setting and assumptions of TE-CDE to other related work.

\begin{table*}[!h]
\centering
\caption{Comparison of problem setting and assumptions. TE-CDE allows irregularly observed data and treatment plans defined in continuous time. Its assumptions are the continuous-time generalization of the standard assumptions in causal inference. The notations are defined in Section 3.}
\vspace{0.1in}
\begin{tabular}{@{}llll@{}}
\toprule
       & Observation time                              & Treatment Plan                              & Overlap                            \\ \midrule
Static setting & $1$                                           & $\{0, 1\}$                                  & $P(A|X)>0$                         \\
CRN/RMSN    & $1,2,\ldots, t \in \mathbb{N}^+$              & $\{t+1, \ldots, t+k\} \rightarrow \{0, 1\}$ & $P(A_{t+1} | \mathcal{F}_{i, t})>0$  \\
TE-CDE & $t_{i,0}, \ldots, t_{i, m_i}\in \mathbb{R}^+$ & $[t, t'] \rightarrow \{0, 1\}$              & $\lambda(t, \mathcal{F}_{i, t})>0$ \\ \bottomrule
\end{tabular}
\label{contrast}
\end{table*}

We argue for modeling the underlying continuous-time processes that give rise to the discrete observational data, which may itself be highly irregular. We contrast this approach with discrete-time methods that use a common discretization for all time series and are forced to interpolate and impute before model fitting. These methods also differ by how they adjust for confounding and for differences in covariate distributions in different treatment regimes. Marginal Structural Models (MSMs) are linear in treatment and covariate effect, and create a pseudo-population using inverse probability of treatment weighting, such that the probability of treatment does not depend on the time-varying confounders and thus effectively controlling for confounding bias \cite{robins2000marginal}. \citet{lim2018forecasting} proposed a semi-parametric alternative to MSMs using recurrent neural networks to estimate propensity weights. The Counterfactual Recurrent Network (CRN, \citealp{bica2020estimating}) uses a similar architecture but instead uses adversarial training to balance differences in covariate distributions in different treatment regimes. However, both assume data to be regularly sampled and fully observed at all time points, which is unrealistic in practice.

Gaussian process-based approaches such as \citet{schulam2017reliable} are applicable to longitudinal data and take a continuous-time approach but in contrast, make strong assumptions about the model structure that is dependent on a particular application and prior knowledge of the form of the processes involved. Closer to the proposed approach, neural ordinary differential equations (ODE, \citealp{chen2018neural,rubanova2019latent}) and extensions \cite{kidger2020neural,morrill2021neural} have been considered for modeling irregular time series data. However, neural ODE type methods are conventional time series models, which do not account for issues such as time-dependent confounding. In the context of intervention modeling, \citet{gwak2020neural} proposed to use separate ODEs for interventions and outcome processes. 
However, they did so for systems with deterministic dynamics without integrating time-varying covariates and without addressing confounding. As a result, their approach is not applicable to treatment effect estimation in healthcare. Related is also \citet{bellot2021policy} that proposed to model treatment effects in continuous time in the context of synthetic controls; however, contrasting our setting where there could be interventions over time, they only consider a single intervention at a particular time point and the approach is not applicable more generally to address multiple treatments.

\section{Problem Formulation}\label{sec:formulation}
We consider $n$ i.i.d. individuals over a study period $[0, T ]$. Each individual is represented by a $d$-dimensional path $\mathbf X :[0,T]\rightarrow \mathbb R^d$, that defines the trajectory of patient covariates over time (and can include static covariates defined to be constant over time), a treatment process $A:[0,T]\rightarrow \{0,1\}$ is a discrete path indicating treatment at each time $t \in [0,T]$, i.e. $A_t=a$, where $a \in \{0,1\}$
and a counting process $N:[0,T]\rightarrow \mathbb N$ to denote the treatment assignment pattern of a single treatment over time, e.g. the number of treatments administered up to a given time\footnote{The definition can be generalized if multiple treatment types are considered, where $N$ (and $A$) can be a multivariate process, each element of $N$ counts a type of treatment assignment and would have a corresponding multivariate intensity process. 
For simplicity, our exposition considers a single treatment only. }. 
These processes are assumed to control or modulate an outcome of interest $Y :[0,T]\rightarrow \mathbb R$, e.g. the tumor size of cancer patients over time, and we will distinguish between potential outcomes of $Y$, denoted $Y(A = a)$ or $Y(a)$ for simplicity, to define the potential outcome trajectory of patient $i$ had it been given a treatment path defined by $A = a$.

In the context of electronic health records (EHRs) and most practical applications, the latent paths $\mathbf X$ are only partially-observed through $m$ irregular observations, $\left\{(t_{0}, \mathbf X_{t_0}),(t_{1}, \mathbf X_{t_1}), \dots, (t_{m}, \mathbf X_{t_m})\right\}$, with each $t_{j} \in \mathbb R$ the timestamp of the observation $\mathbf X_{t_j} \in \mathbb R^d$.

To avoid notation clutter, we use the time subscript to refer to function evaluation. The same observations apply to paths $A$ and $Y$. The case where each $i$-th patient observation sequence has its own $m_i$ irregular time stamps $t_{i,0},\dots, t_{i,m_i}$, thus  
differences in sampling intensity within a patient's trajectory and between different patients can be considered without modification of any part of the exposition. Indeed, analyzing time series data with such a complex pattern of observation is the central motivation of this work. Let $\mathcal F_{t}$ denote the filtration that is generated by all the observable events for a given individual up to time $t$, including observations of $\mathbf X_s, A_s$ and $Y_{s}$ for $s \leq t$.

Our goal is to derive unbiased estimates of the potential outcomes at a given time $t'$: $\mathbb E[Y_{t'}(A=a)|\mathcal F_{t}]$, for any value of time in the future $t'>t$, hypothesized discrete treatment path $A:[t, t']\rightarrow \mathbb \{0,1\}$ with values $a$, given past observations up to time $t$, $\mathcal F_{t}$. However, with observational data only one of these potential outcomes trajectories is observed for each unit depending on the treatment assignment. We refer to the unobserved potential outcomes as counterfactuals. 

\begin{figure*}[t]
    \centering
    \includegraphics[width=0.75\textwidth]{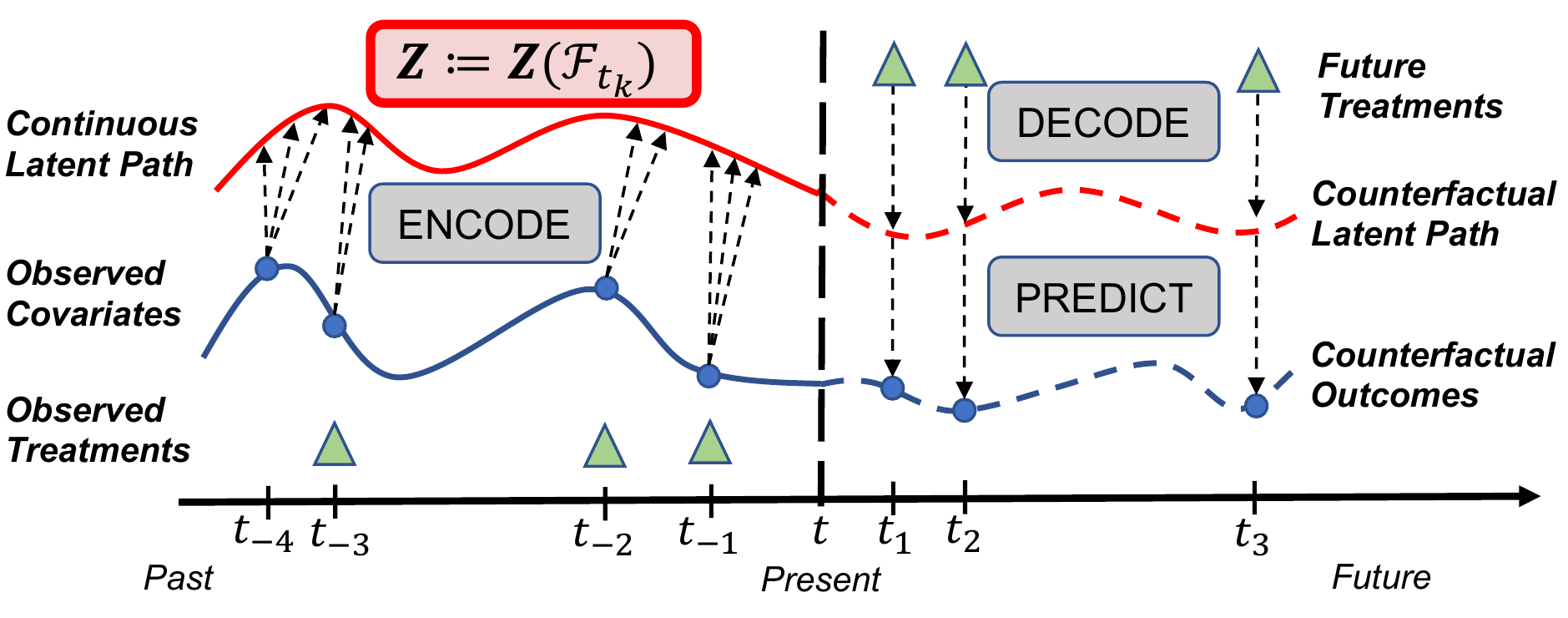}
    \caption{An illustration of TE-CDE. We learn a continuous latent path $\mathbf{Z}_{t}$ as the solution to a CDE by encoding historical observations. At future time points, we decode (hypothetical) future treatments to determine the latent state and use this to predict counterfactual outcomes.}
    \label{fig:method}
\end{figure*}

Potential outcomes processes are identifiable with respect to the filtration generated by the observed data under the following three assumptions. These three assumptions are the standard causal inference assumptions.

\textbf{Assumption 1} (Consistency). \textit{For an observed treatment process $A = a$, the potential outcome under this treatment trajectory is the same as the factual outcome $Y(a)=Y$.}

\textbf{Assumption 2} (Overlap). \textit{The intensity process $\lambda(t|\mathcal F_t)$
is not deterministic given any filtration $\mathcal F_{t}$ and time point $t\in[0,T]$, i.e.
\begin{align}
    0 <\lambda(t|\mathcal F_t)=\lim_{\delta t\rightarrow 0}\frac{P(A_{t+\delta t} - A_t = 1 | \mathcal F_{t})}{\delta t} < 1.
\end{align}}

Overlap means that there is some positive probability of treatment assignment at any point along a patient's trajectory over the time interval. It can be understood as a direct extension to the more familiar overlap assumption in the static context, $0<P(\text{Treatment} = 1 | \mathbf x) < 1$.

The last assumption extends unconfoundedness, or strong ignorability given a patient's trajectory, to ensure that it is sufficient to condition on the observed trajectory up to time $t$ to block all backdoor paths, i.e. spurious correlation not part of the direct causal effect of interest, to the potential outcome at any time in the future. Similar to Assumption 2, unconfoundedness has previously been extended to the continuous-time domain for stochastic processes by \citet{lok2008statistical,saarela2016flexible,ryalen2019additive}. 

\textbf{Assumption 3} (Continuous-time sequential randomization). \textit{The intensity process $\lambda(t|\mathcal F_{t})$ with respect to the filtration $\mathcal F_{t}$ is equal to the intensity process generated by the filtration $\mathcal F_{t}\cup \{\sigma(Y_{s}):s>t\}$ that includes the $\sigma$-algebras generated by future outcomes $\{\sigma(Y_{s}):s>t\}$.}

It is worth mentioning that the intensity process plays the same role as propensity scores in discrete-time models \cite{robins1997}, modeling the switching of the treatment process. Assumption 3 can thus be thought of as formalizing sequential randomization in the continuous-time model by stating that the intensity process does not depend on future potential outcomes, i.e. the current information is enough to estimate counterfactuals in the future without bias.

\section{Treatment Effect Controlled Differential Equation}\label{sec:method}
TE-CDE frames the latent trajectory of a patient's state, as a response to a controlled differential equation (CDE), driven by covariate, treatment, and outcome processes  (Fig. \ref{fig:method}), which to the best of our knowledge is the first to do so.

This formulation using a CDE permits to account for  information available at $t>0$ (rather than just initial value $t=0$).
In particular, neural controlled differential equations \cite{kidger2020neural,morrill2021neural} allow incoming information to modulate the dynamics. This ability is natural in a clinical setting, as not only can we model the continuous-time latent state evolution of a patient trajectory, but also we account for incoming data (e.g. treatment changes) that modulate the dynamics of the system.

We now present key components needed to facilitate the modeling of \textit{counterfactual} outcomes in \textit{continuous} time. Additional properties of TE-CDE are discussed in Appendix \ref{appendixC}.
The key components are as follows:\\
(1) TE-CDE's encoder learns a representation that is defined continuously in time (i.e. a continuous latent path), rather than only at discrete time steps. \\
(2) The latent path trajectory evolves as a response of a Neural Controlled Differential Equation (CDE).\\
(3) Decoding and prediction are in continuous-time. \\
(4) TE-CDE uses domain adversarial training to learn a representation that adjusts for time-dependent confounding and hence is suitable for causal estimation.

\textbf{Encoding the latent path Z.} 
TE-CDE's encoder ingests historical observations $\mathcal F_{t}$ up to time $t$ and learns a latent path $\mathbf{Z}: [t_0, t] \rightarrow \mathbb{R}^l$ continuously over time that will be designed to be both predictive of the factual outcomes and agnostic of the observed assigned treatment. An explicit continuous-time representation allows us to process measurements with arbitrary observation patterns. 
We assume the initial state of the path $\mathbf Z_{t_0}$ to be parameterized by a neural network $g_{\eta}:\mathbb R^{d+1+1} \rightarrow \mathbb R^{l}$ embeds the initial outcome, covariate and treatment observations into a $l$-dimensional latent state which can be expressed as the solution to a CDE,
\begin{align}
\label{cde}
     \mathbf Z_{t_0} = g_\eta(\mathbf X_{t_0}, A_{t_0}, Y_{t_0}),\\
     \mathbf Z_t = \mathbf Z_{t_0} + \int_{t_0}^{t} f_{\theta}(\mathbf Z_s)\frac{d[\mathbf X_s,A_s,Y_s]}{ds}ds,
\end{align}
for $t \in (t_0, T]$ which denotes the present time, up to which observations of all processes are available. The dynamics of potential outcomes when controlled by the covariate and treatment process take the form of a CDE \cite{lyons2007differential}. Hence, the solution $\mathbf Z$ is said to be the response of a Neural CDE  \cite{kidger2020neural} driven or controlled by the covariate, treatment and outcome processes (concatenated into a vector $[\mathbf X_t,A_t,Y_t]\in\mathbb R^{d+1+1}$). In this sense, Neural CDEs are a family of continuous-time models that explicitly define the latent vector field $f_{\theta}: \mathbb R^{l} \rightarrow \mathbb R^{(d+1+1) \times l}$ by a neural network parameterized by $\theta$, and the dynamics are modulated by the values of an auxiliary path over time.

We computationally obtain the latent path up to $t$ from $\mathcal F_{t}$ by solving the above initial value problem (IVP): $\forall s \in [t_0, t]$,
\begin{align}
\label{eq:cde_solve}
    \mathbf{z}_{s} = 
    \text{ODESolve}(f_{\theta}, \mathbf{Z}_{t_0}, \mathbf X_{\le t}, A_{\le t}, Y_{\le t}),
\end{align}
where ODESolve denotes a numerical ODE solver as proposed by \citet{kidger2020neural}. 

In practice, we have access to observations at certain (irregular) time points. Thus, we define an interpolation of the data with piece-wise continuous derivatives that serves as an approximation of the underlying paths\footnote{We note that since there may be discontinuities in $A_{\le t}$ (e.g. treatment is applied in discrete stages) we can inform the solver about the jumps between pieces so that its integration steps may align with them. This can be achieved using the \texttt{jumpt} argument in the ODE solver \cite{chen2018neural,kidger2020neural}.}.

\textbf{Decoding and prediction.} 
After the encoder processes all the observations up to the present time $t$, TE-CDE starts to decode and predict the potential outcomes up to some time $t' > t$ in the future for a hypothetical treatment schedule defined by the user. At this point, the latent path $\mathbf Z$ potentially changes as a result of the chosen treatment schedule, which can similarly be formalized using a second controlled differential equation such that,
\begin{align}
\label{cde-decode}
     \mathbf Z_{t'} = \mathbf Z_{t} + \int_{t}^{t'} f_{\phi}(\mathbf Z_s)\frac{dA_s}{ds}ds,
\end{align}
where $t'$ denotes a desired time horizon, $\mathbf Z_{t}$ is the latent state of $\mathbf Z$ at time $t$ which encodes the patient's history, and $A_s$ represents the hypothetical treatment schedule for $t<s<t'$. $f_\phi: \mathbb{R}^{l+1} \rightarrow \mathbb{R}^l$ is a feed-forward neural network with trainable weights $\phi$.
As before, the decoded path can be obtained by solving the IVP:
\begin{align}
    \mathbf{Z}_{s} = 
    \text{ODESolve}(f_{\phi}, \mathbf{Z}_{t}, A_{t\leq t'}).
\end{align}

\textbf{Domain adversarial training for counterfactual estimation}.
The covariates $\mathbf X$ are time-dependent confounders, which can increase variance in the estimation of counterfactuals if the treatment distribution is not properly balanced given a patient's trajectory \cite{mansournia2017handling}. While unbiased by Assumption 3, counterfactual estimates may have lower variance given patient trajectories frequently observed in the data but higher variance for infrequently observed patient trajectories with consequences for performance generalization of the treatment effect as demonstrated by \citet{shalit2017estimating}. 
To mitigate this confounding bias, we ensure the latent representation $\mathbf Z_{t}$ is \textit{not} predictive of the observed treatment assignment pattern \cite{shalit2017estimating,bica2020estimating} which effectively induces representations that are balanced with respect to treatment assignment over time. The treatment invariance breaks the association between time-dependent confounders $X_t$ and current treatment $A_t$.

At each time $t$, the $j$ different treatments $A \in \{A_1 , \dots ,A_j \}$ represent our domains. We then require at each timestep $t$, that the latent path $Z_t$ be invariant across treatments options: $P(\mathbf Z_t | A_t = 0) = P(\mathbf Z_t | A_t = 1)$ and more generally equal across any two values in the domain of treatment options. In this context, distributions of the latent state differ across treatment groups if a classifier trained as a function $\mathbf Z_t$ to predict treatment assignment accurately separates the two groups. Such representations are called \textit{balancing representation} as it balances the probability of the predicted treatment process $p(A_{t} = 1 | \mathbf{Z}_{t}) = 0.5$, i.e. minimizing the distributional variance between treatment groups in the representation space \cite{johansson2020generalization}. 

We use two neural networks $h_{\nu}:\mathbb R^{l} \rightarrow \mathbb R^{d}$ and $h_{a}:\mathbb R^{l} \rightarrow [0, 1]$ to predict the outcome and treatment: $\forall s \in [t, t']$
\begin{align}
    &\hat y_{s} = h_{\nu}(\mathbf{z}_{s}),\\
    &\hat p_{s}:= \hat p(a_{s} = 1) = h_{\alpha}(\mathbf{z}_{s}).
\end{align}
Suppose there are $k \ge 1$ observations in the time window $[t, t']$ with observation times $(t_1, \ldots, t_k)$. The mean square error (MSE) of outcome prediction is defined as $\mathcal{L}^{(y)}= \frac{1}{k}\sum_{j=1}^k \left(y_{t_j} - \hat{y}_{t_j}\right)^2$. The cross entropy loss of treatment prediction is defined as: $\mathcal{L}^{(a)}= - \frac{1}{k}\sum_{j=1}^k a_{t_j} \log\left(\hat p_{t_j}\right) + (1-a_{t_j}) \log\left(1 - \hat p_{t_j}\right)$.

We minimize the following loss function, enforcing simultaneous outcome prediction and balanced representations:
\begin{align}\label{eq:loss_fn}
    \mathcal{L}=\frac{1}{n}\sum_{i=1}^{n} \mathcal{L}^{(y)}_{i}-\mu \mathcal{L}^{(a)}_{i},
\end{align}
where $\mu > 0$ is a hyper-parameter controlling the trade-off between treatment and outcome prediction. Note that the minus sign before $\mathcal{L}^{(a)}$ would effectively \textit{maximize} the treatment prediction loss and ensure that $z_{t}$ is not predictive of treatment assignment $A_t$. This leads to balancing representations, which remove bias introduced by time-dependent confounders and allow for reliable counterfactual estimates.

\textbf{\textit{Remarks on invariant representations}}. As shown by \citet{johansson2019support}, invertible transformations ($\phi$) are necessary for consistency of domain invariant representations ($\mathbf Z$). We include for completeness that if $\phi$ is non-invertible there is information loss, which leads to unobservable error ($\eta$).  Thus, we desire an invertible $\phi$, which ensures $\eta=0$. 
This highlights an important \emph{strength of TE-CDE}, where by properties of ODEs/CDEs \cite{invertzhang2020}, the representations from TE-CDE have \emph{guaranteed invertibility}, since integration backward in time is always possible or we can alternatively integrate: $-f_{\phi}(\mathbf Z_s)$ .

\textbf{Intensity of sampling}. 
It is well-known for EHR data that sampling frequency and observations (or lack thereof)
carry information about the patient's health status \cite{alaa2017learning}. In such cases, we can replace each observed tuple $(x_{t_j}, a_{t_j}, y_{t_j})$ with $(x_{t_j}, a_{t_j}, y_{t_j}, c_{t_j})$ where $c_{t_j}\in\mathbb R^{d+1+1}$ counts the number of times each one of the dimensions of $X$, $A$ and $Y$ have been observed up to time $t_j$. The extended tuple is fed into the encoder to inform it about the sampling.

\section{Experiments}\label{sec:experiments}
In this section, we validate the ability of TE-CDE to estimate counterfactual outcomes from irregularly sampled observational data.
Since counterfactual outcomes are not known for real-world data, it is necessary to use synthetic or semi-synthetic data for empirical evaluation. 
First, we describe a simulation environment based on a Pharmacokinetic-Pharmacodynamic (PK-PD) model of lung cancer tumor growth \cite{geng2017prediction}, which allows counterfactuals to be calculated at any time point for arbitrary treatment plans.
Furthermore, we introduce a continuous-time observation process based on Hawkes processes. The controllable nature of the observation process allows us to simulate irregularly sampled observational data for a range of different observation process parameterizations, which are motivated by common healthcare scenarios. 


\subsection{Modeling tumor growth under general observation patterns}\label{sec:data_env}

\textbf{Tumor growth dynamics.} 
We use a well-established bio-mathematical PK-PD model for tumor growth in lung cancer patients that includes the effects of chemotherapy and radiotherapy \cite{geng2017prediction}. The PK-PD model is representative of the true underlying physiological process with responses to interventions. Hence, results using the model should be closely representative of reality. Additionally, the same underlying model was also used by \citet{lim2018forecasting} and \citet{bica2020estimating}. We briefly describe it below and refer the reader to Appendix \ref{appendixB} for more details.
The tumor volume at time $t$ after diagnosis is modeled as follows:
\begin{align}
\begin{split}
\frac{dV(t)}{dt} &=\bigg(\underbrace{\rho \log \left(\frac{K}{V(t)}\right)}_{\text {Tumor growth }}-\underbrace{\beta_{c} C(t)}_{\text {Chemotherapy }}\\
&-\underbrace{\left(\alpha_{r} d(t)+\beta_{r} d(t)^{2}\right)}_{\text {Radiotherapy }}+\underbrace{e_{t}}_{\text {Noise }} \bigg) V(t),
\end{split}
\label{eq:pkpd}
\end{align}
where chemotherapy concentration $C(t)$ and radiotherapy dose $d(t)$ are defined by their own equations (see Appendix \ref{appendixB1}),
$K, \beta_{c}, \alpha_{r}, \beta_{r}$ are effect parameters, and $e_{t}$ accounts for randomness in tumor growth \cite{geng2017prediction}.

We consider four treatment options: no treatment, chemotherapy, radiotherapy, and combined chemotherapy and radiotherapy.  The assignment of chemotherapy and radiotherapy are modeled as Bernoulli random variables with probabilities $p_{c}$ and $p_{r}$, respectively, that depend on tumor diameter\footnote{Note we assume tumors are perfectly spherical to enable conversion between tumor volume and diameter.} as follows:
$p_{c}(t)=\sigma\left(\frac{\gamma_{c}}{D_{\max }}\left(\bar{D}(t)-\theta_{c}\right)\right)$,
$p_{r}(t)=\sigma\left(\frac{\gamma_{r}}{D_{\max }}\left(\bar{D}(t)-\theta_{r}\right)\right)$,
where $D_{\max}=13 \textnormal{cm}$ is the maximum tumor diameter, $\theta_{c}=\theta_{r}=D_{\max}/2$ and $\bar{D}(t)$ is the average tumor diameter. 

The degree of \textit{time-dependent confounding} is controlled by $\gamma_{c}$ and $\gamma_{r}$, where increasing $\gamma_{\{c,r\}}$ increases the probability that observational treatment assignment is based on tumor diameter (For more details see Appendix \ref{appendixB1}).

\begin{figure*}[!th]
  \centering
  \subfigure[$\kappa$=1]{\includegraphics[width=0.25\textwidth]{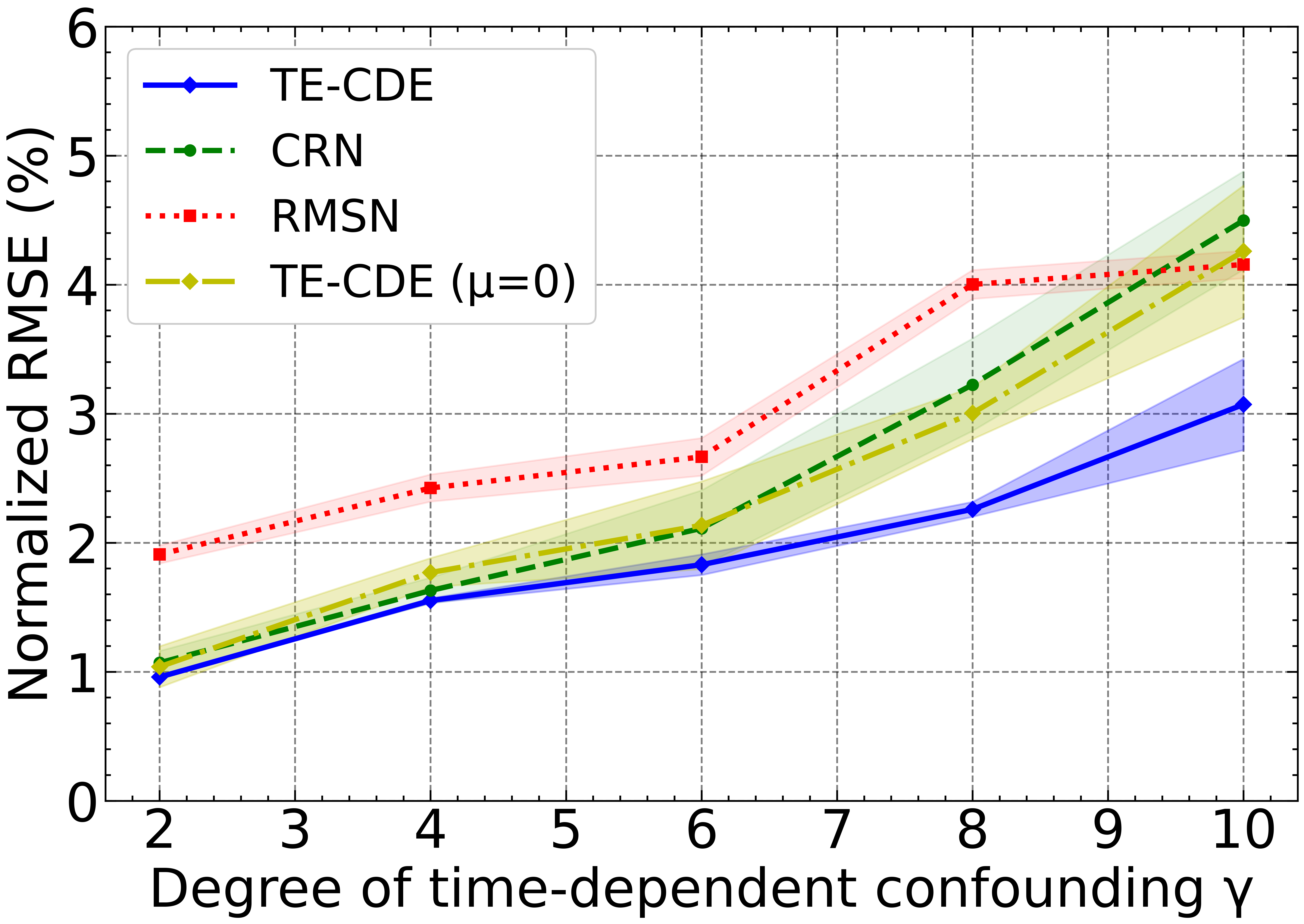}}\quad\quad
  \subfigure[$\kappa$=5]{\includegraphics[width=0.25\textwidth]{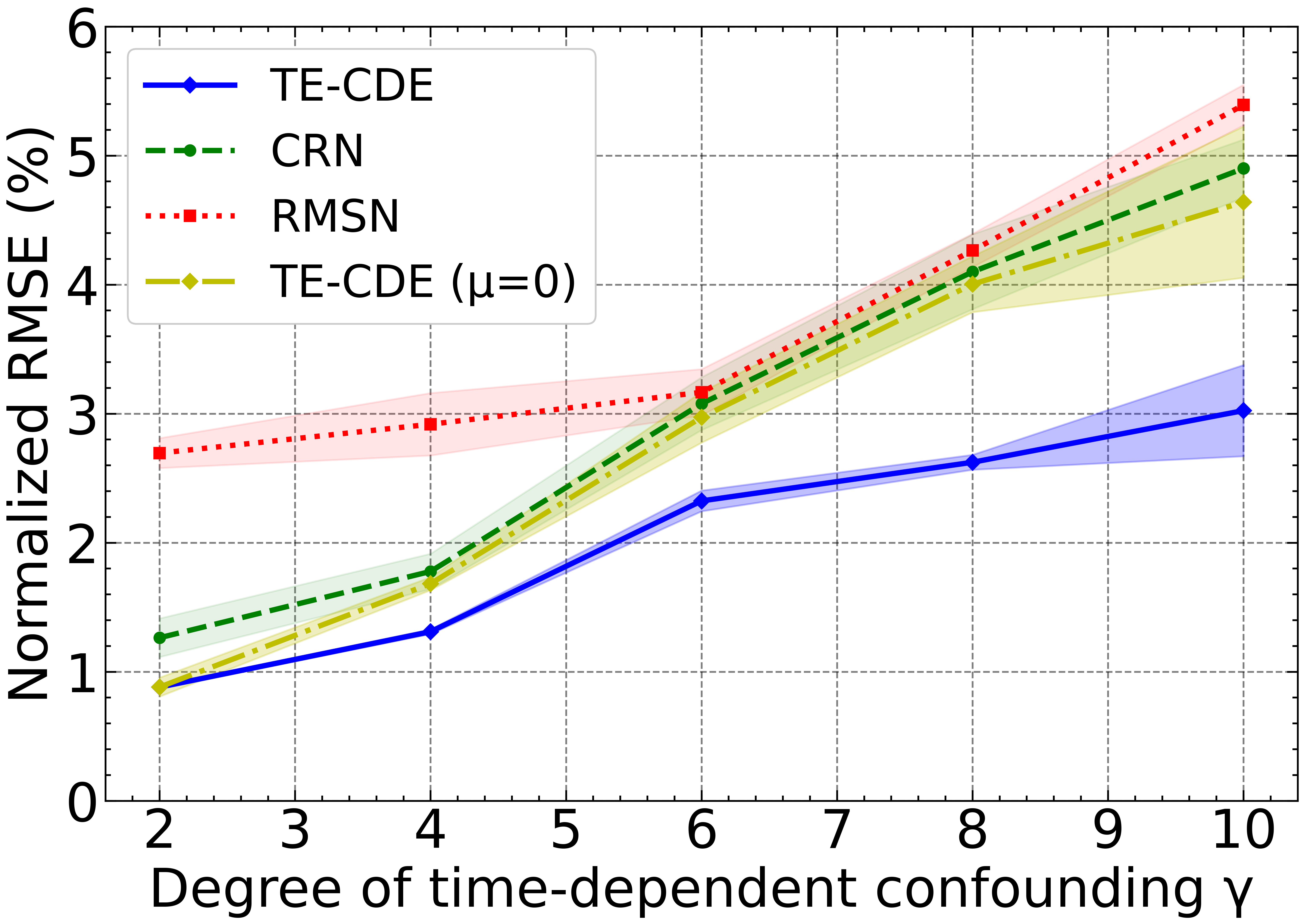}}\quad\quad
  \subfigure[$\kappa$=10]{\includegraphics[width=0.25\textwidth]{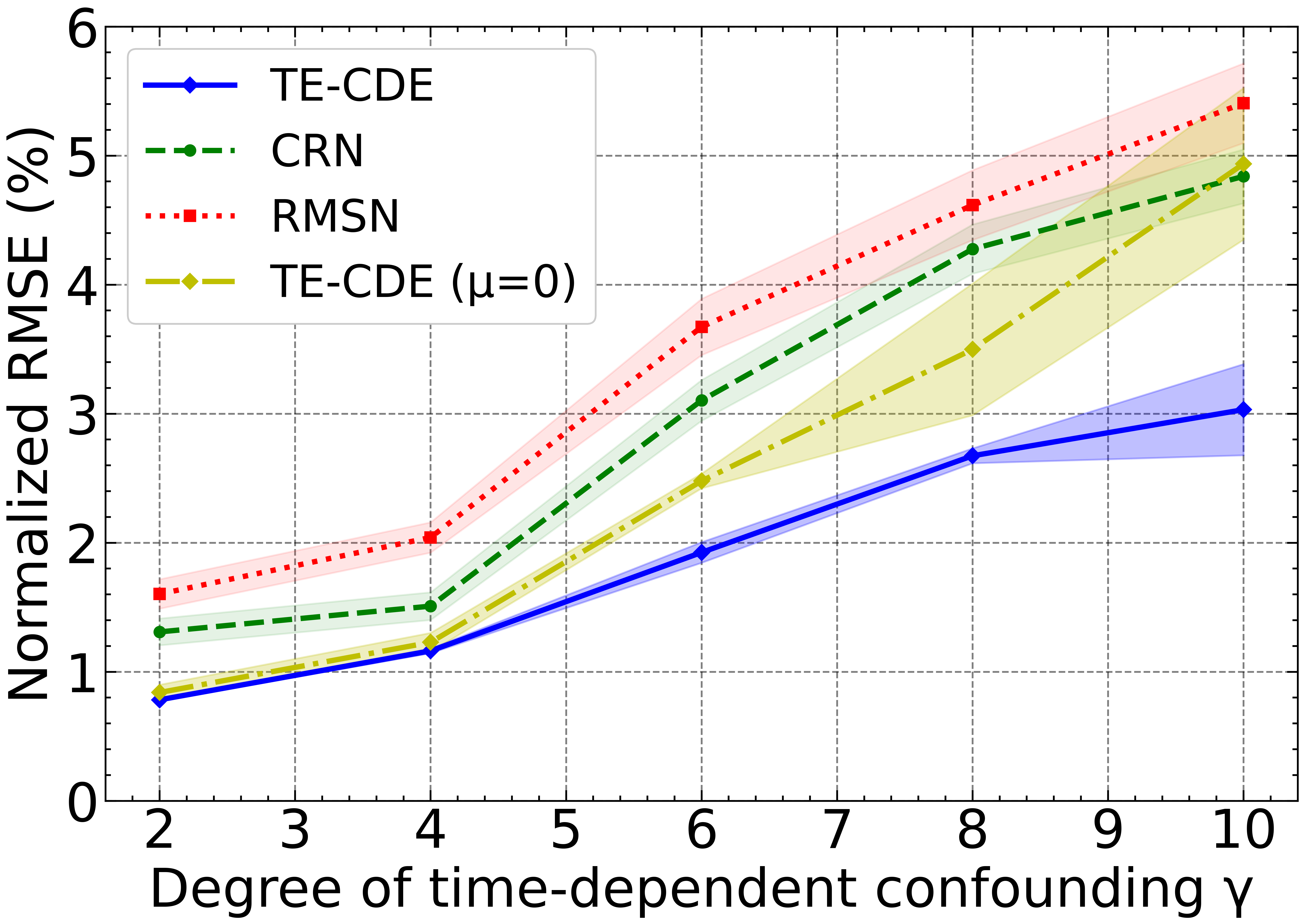}}
  \caption{Counterfactual predictions for varying time-varying confounding $\gamma$. Additionally, various values of $\kappa$ are illustrated which controls the intensity of sampling varying over a trajectory based on clinical stage.}
   \label{fig:sweep}
\end{figure*}

\textbf{Observation process.} 
As discussed in Section \ref{sec:intro}, in real-world clinical settings, patients are rarely observed at fixed, regular time intervals. Instead, they are observed \textit{irregularly}, with observations often a consequence of clinical factors, e.g. severity of illness, treatment regimen, or medical policy. 

To simulate such nuances, we augment the simulation environment by modeling the patient observation process with a Hawkes process \cite{hawkes1971spectra,hawkes1974cluster}. 
A Hawkes process is a flexible point process with temporal dependencies. Indeed, since clinicians are not memoryless and the times when they make observations often depend on past observations, the Hawkes process appears a sensible parameterization of such an observation process. This is especially true since it captures time-varying sampling intensity that depends on both patient history and clinical state.

These properties have been leveraged by both \citet{bao2017hawkes} and \citet{alaa2017learning}, who used a Hawkes process to model different types of observational healthcare data, further justifying its applicability to healthcare and its use in our simulations. In addition, the success of both methods applying Hawkes processes to EHR data (which is often high-dimensional data), underlines the applicability to high-dimensional scenarios. From an experimental perspective, the Hawkes process is readily parameterized to simulate different clinical observation scenarios/regimes (i.e. a test bed to simulate different clinical scenarios).

Formally, the observation process $\{t_m\}_{m \in \mathbb{N}}$ is modeled as a univariate Hawkes process with a self-exciting intensity function and exponential kernel function \cite{lee2016hawkes}:
\begin{equation}
\begin{split}
\lambda(t,s_t=i) = \lambda_{i}^{o} + \sum_{\tau < t_m < t}e^{-2 (t-t_m)},  \\
\lambda_i^{o} = 0.01\kappa^{i},
\end{split}
\label{eq2}
\end{equation}
where $s_t$ represents the clinical state of the patient at time $t$ and $\tau$ represents the time of the previous change of state, i.e. $\tau = \max \{t^{\prime}<t: s_t^{\prime} = s_t,\ s_{t^-}^{\prime} \neq s_t\}$.
We define the clinical state using the American Joint Committee on Cancer staging system \cite{tnm}. We convert the clinical stages $\{S1A,S1B,S2,S3\}$ to stages $s_t=\{0,1,2,3\}$ (see Appendix \ref{appendixB2} for further details).

While in state $s_t=i$, the base observation intensity is given by $\lambda_i^{o} = 0.01\kappa^{i}$, where $\kappa \geq 1$ controls the change of base sampling intensity between cancer stages, resulting in a higher base sampling intensity for patients with more advanced cancer, emulating real-world observation patterns. 

Note that further control over the sampling process is possible, either by introducing additional states $s_t$ or varying $\kappa$ based on, for example, treatment.
Examples of observation trajectories can be found in Appendix \ref{appendixB2}.

\textbf{Benchmarks.} We compare TE-CDE \footnote{https://github.com/seedatnabeel/TE-CDE}\textsuperscript{,}\footnote{https://github.com/vanderschaarlab/mlforhealthlabpub/\newline tree/main/alg/TE-CDE}  with state-of-the-art methods for counterfactual estimation over time CRN \cite{bica2020estimating} and RMSN \cite{lim2018forecasting} in different irregular sampling scenarios. Both CRN and RMSN rely on the assumption of regularly sampled data. Thus, for the irregular setting of interest, we divide the timeline equally, interpolate and impute the ``un-observed'' observations. We also evaluate a Gaussian Process (GP) based model for continuous-time, similar to \citet{schulam2017reliable}. However, the model performance is poor for larger values of time-dependent confounding, and the results are included together with additional experiments in Appendix \ref{appendixE}. For domain adversarial training, we use the standard procedure \cite{ganin2016}, with an initial $\mu=0$ that follows an exponentially increasing schedule per epoch of training for the range $[0,1]$.
In addition, we assess the impact of the adversarial training procedure in TE-CDE by training a version of our method \emph{without} domain adversarial training, i.e. constant $\mu=0$ in the loss function (Eq. \ref{eq:loss_fn}).

\textbf{Experimental details.} 
Implementation details, including hyper-parameters, can be found in Appendix \ref{appendixD}.
Unless otherwise stated, each experiment is run with 10,000 patients for training, 1,000 for validation and 10,000 for testing.

\subsection{Impact of time-dependent confounding across varying sampling intensities}\label{sec:exp-time}
A key challenge when learning from longitudinal data is accounting for bias introduced by time-dependent confounding.
Therefore, it is essential to assess the impact of time-dependent confounding on counterfactual estimation. 
We measure performance via normalized RMSE, where the RMSE is normalized by the maximum tumor volume $V_{\max} = 1150 \textnormal{cm}^3$.
As discussed, time-dependent confounding is controlled by parameters $\gamma_{c}$, $\gamma_{r}$ in the treatment assignment policies. 
We evaluate the benchmarks under increasing degrees of time-dependent confounding by setting $\gamma_{c}=\gamma_{r}=\gamma=\{2,4,6,8,10\}$.

As previously discussed, a patient's condition often affects the frequency of observation. The state-based variability in sampling intensity based on the clinical state is controlled by $\kappa$ in the simulation environment. We repeat the experiment for multiple values of the scaling factor $\kappa=\{1,5,10\}$.

Figure \ref{fig:sweep} shows the results for counterfactual estimation for different levels of sampling intensity $\kappa$. 
As expected, the performance of all models degrades with increasing time-dependent confounding.
However, TE-CDE achieves the lowest counterfactual estimation RMSE for all values of sampling intensity $\kappa$ and across all values of time-dependent confounding $\gamma$. 
The divergence is most pronounced for increasing $\gamma$, with $\gamma=10$ leading to a $36\%$ decrease in RMSE for TE-CDE compared to CRN, the next best performing method.
The superior performance of TE-CDE in all settings highlights the benefit of the continuous-time approach adopted compared to RNN-based approaches.

Comparing the RNN-based models, CRN outperforms RMSN, matching the conclusions in the regularly sampled setting reported in \citet{bica2020estimating}. Overall, however, the results highlight the limitation of RNN-based models in the irregularly sampled setting.

Finally, we characterize the value of domain adversarial training in TE-CDE by comparing it to the case when $\mu=0$ (i.e. no domain adversarial training). TE-CDE ($\mu=0$) suffers significant performance degradation with a higher RMSE in all scenarios that grows as the degree of time-dependent confounding increases. This clearly demonstrates the practical benefit of the adversarial training approach to learning balancing representations.

\vspace{-1mm}
\subsection{Treatment-conditioned sampling}

In the previous section, we consider the realistic scenario where the severity of the patients' condition governs the intensity of the observation process, i.e. sicker patients with higher stages are observed more frequently. 
In addition, the treatment regimen itself will often also influence the observation pattern of the patient, i.e. patients undergoing different treatments are monitored differently.

To simulate this phenomenon, we adjust the base sampling intensity $\kappa$ depending on whether the patient is treated or untreated, thereby altering the state-dependent sampling variability between the two treatment groups. We set $\kappa=10$ for treated patients and $\kappa=1$ for untreated patients. We fix the time-dependent confounding as $\gamma=4$.

Consistent with the previous experiment, TE-CDE significantly outperforms the benchmark models (Table \ref{table:results-treatment}). 
There is a divergence in performance between the treated and untreated populations. 
This is largely explained by differences in the severity of the clinical condition between the two populations: the majority of untreated patients ($77\%$) remain in cancer stage S1A (i.e. $\max s_t=0$). 
Due to the reduced state transition, these patients naturally have lower variability in tumor volume over the trajectory, leading to lower error for all methods.  
Overall, 
TE-CDE outperforms all methods globally and for treated and untreated patients.
\begin{table}[ht]
    \centering
        \caption{Normalized RMSE (\%) for counterfactual estimation with $\kappa$ conditioned on treatment, where $\kappa_{treated}=10$ and $\kappa_{untreated}=1$.}
\scalebox{0.95}{
\begin{tabular}{cccc}
    \toprule

Model   &  Overall         & Treated         & Untreated       \\ \midrule \midrule
TE-CDE  & 1.18 $\pm$ 0.05  & 1.56 $\pm$ 0.06 & 0.22 $\pm$ 0.02 \\ \midrule \midrule
CRN     & 1.57 $\pm$ 0.06  & 1.97 $\pm$ 0.05 & 0.64 $\pm$ 0.08 \\ \midrule \midrule
RMSN    &  3.06 $\pm$ 0.09 & 3.12 $\pm$ 0.07 & 2.83 $\pm$ 0.08 \\ \midrule
  \bottomrule
\end{tabular}}
    \label{table:results-treatment}
\end{table} \vspace{-1mm}
\subsection{Forecasting at additional time horizons}\label{sec:exp-future}
We have assessed the ability to estimate counterfactual outcomes at the subsequent observation time determined by the Hawkes process described in Section \ref{sec:data_env}. To further validate our method, we assess counterfactual estimation at subsequent observation times that are further in the future.

\begin{figure}[ht]
    \centering
    \includegraphics[width=0.375\textwidth]{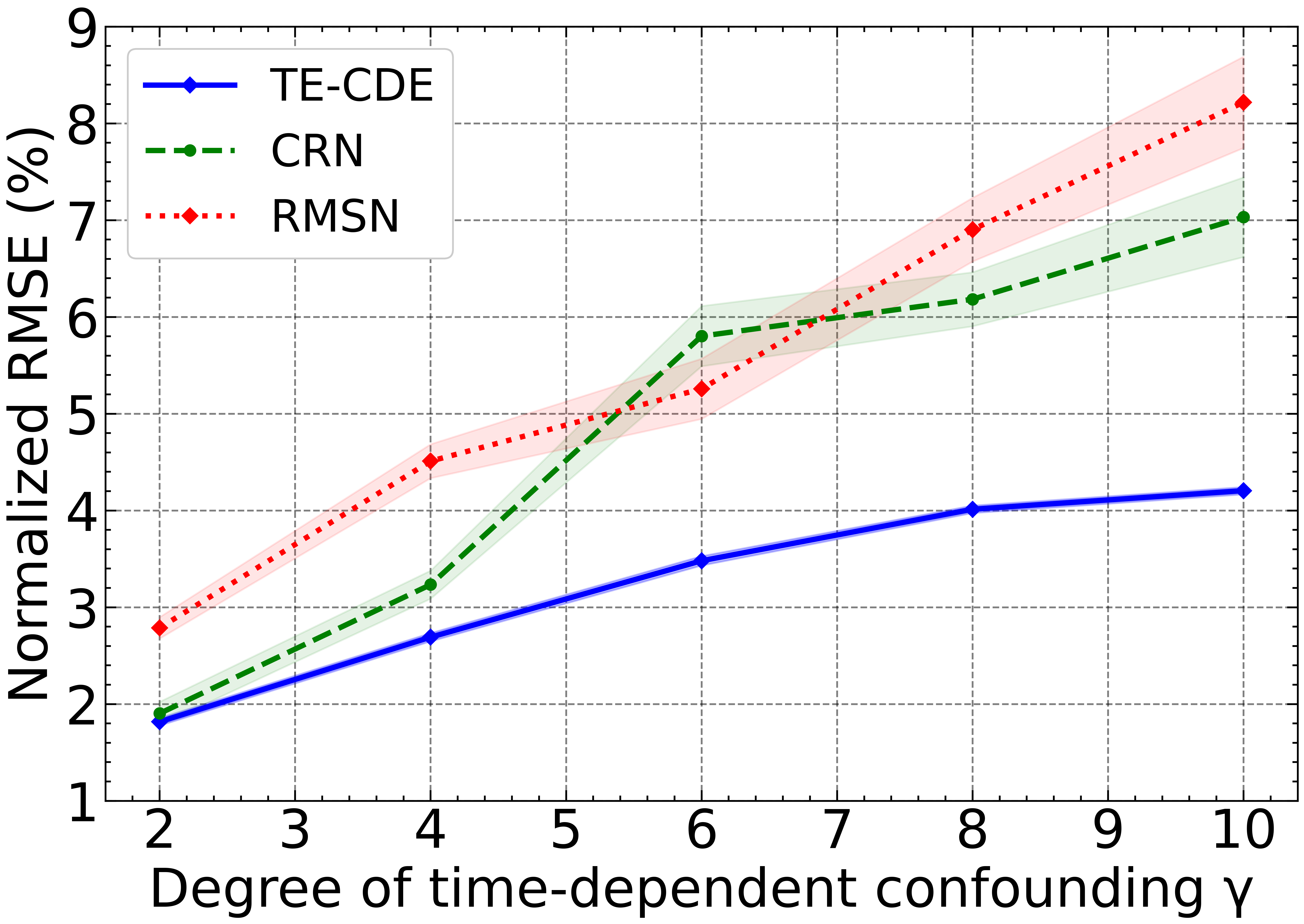}
    \caption{Counterfactual estimation at time $t_{k+n}$ ($n=5$) for varying levels of time-dependent confounding $\gamma$.}
    \label{fig:multistep}
\end{figure}

As an illustrative example, we evaluate counterfactual estimation
at time $t_{k+n}$, i.e. estimate tumor volume $y_{t_{k+n}}$.
Similar to other experiments, we vary the degree of time-dependent confounding $\gamma=\{2,4,6,8,10\}$, fix $\kappa=10$, and set the forecasting horizon $n=5$ (see Appendix \ref{appendix-horizons} for other time horizons). As expected, it is more challenging to estimate counterfactuals further in the future.
As shown in Figure \ref{fig:multistep}, similar trends are observed as the setting of Section \ref{sec:exp-time} (Figure \ref{fig:sweep}). 
TE-CDE outperforms both CRN and RMSN for all $\gamma$, with a greater performance differential as the degree time-dependent confounding increases. For $\gamma=10$, there is a $ 40\%$ reduction in RMSE for TE-CDE.

\subsection{Treatment selection}
To demonstrate how models such as TE-CDE could be used in decision support and the potential impact assisting clinical decision makers, we must assess performance in ways beyond counterfactual estimation.
One such clinically relevant evaluation is whether
the best treatment was selected. 
This is important since reduced error in counterfactual estimation does not necessarily result in improved clinical outcomes.

We define the ``correct'' treatment selection as the treatment that minimizes the tumor volume at time $t_{k+n}$ (i.e. $y_{t_{k+n}}$). 

We adopt the same experimental setup as Section \ref{sec:exp-future} and set the forecasting horizon $n=5$, sampling intensity $\kappa=10$, and vary time-dependent confounding $\gamma=\{2,4,6,8,10\}$.

\begin{figure}[ht]
\centering
    \includegraphics[width=0.375\textwidth]{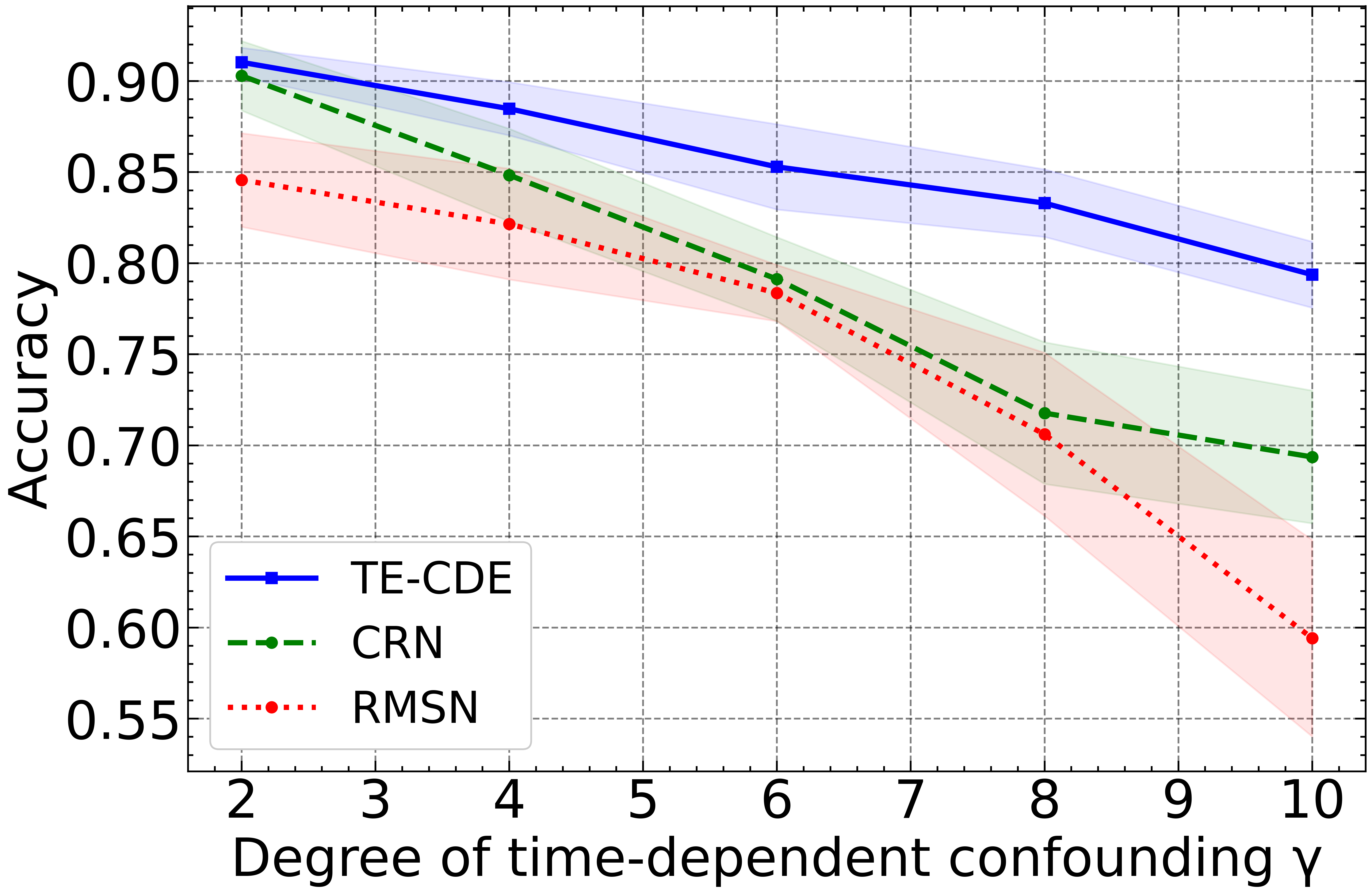}
  \caption{Treatment accuracy under varying degrees of time-dependent confounding $\gamma$.}
  \label{fig:accuracy}
\end{figure}

Figure \ref{fig:accuracy} shows decreasing accuracy for all methods as time-dependent confounding increases. 
However, for all values of $\gamma$, TE-CDE outperforms both CRN and RMSN, more frequently selecting the optimal treatment. Similar to the other experiments, as $\gamma$ increases, the performance gap between TE-CDE and both CRN and RMSN increases, with c. $4\%$ difference in absolute treatment selection accuracy at $\gamma=4$ increasing to c. $10\%$ at $\gamma=10$.
This experiment emphasizes that differences in counterfactual estimation result in meaningful differences in treatment selection accuracy.

\subsection{Additional experiments}
We perform a number of additional experiments to further validate TE-CDE. In Appendix \ref{appendix-uncertain}, we show how uncertainty estimates can be obtained from TE-CDE and then used to rank counterfactual estimates such that uncertain samples can be deferred to clinicians to improve outcomes. 
Appendix \ref{appendix-efficiency} compares the data efficiency of the different methods, which is useful in clinical settings with limited labeled data. 
In Appendix \ref{appendix-latent}, we explore the latent representation of TE-CDE over time to: (1) highlight that the latent states $z_{t}$ learned by TE-CDE indeed are treatment-invariant representations and (2) investigate clinical insights that can be ascertained from the latent representations.

\section{Conclusion}\label{sec:conclusion}
State-of-the-art methods for counterfactual estimation are predicated on the assumption of regular and evenly spaced data sampling. However, real-world clinical time series are often irregular. 
To address this challenge,
we introduce TE-CDE, a model that learns to perform counterfactual estimation in continuous time from irregularly sampled observational data with time-dependent confounding. 
Additionally, we propose a controlled simulation environment for medically realistic irregularly sampled time series. In experiments in a variety of irregular settings, we demonstrate that TE-CDE provides improvements over current state-of-the-art methods. 

Counterfactual estimation has the potential to assist clinicians with “what-if” decision-making. However, when deploying such models in healthcare settings, there are risks, e.g. inaccurate predictions. 
Furthermore, trade-offs between outcomes and possible treatment side effects are not accounted for by such models. To mitigate possible adverse effects, counterfactual estimates should be part of a “human-in-the-loop" paradigm, allowing experts to complement predictions with domain knowledge to improve patient outcomes.

We also note that while some aspects of irregularly sampled data are naturally addressed through our formulation, this work is simply a step in the right direction and our proposed solution only partially addresses the complexities of irregular sampling. In particular more is needed to address a number of aspects, such as informative sampling, that are widely prevalent in healthcare. We hope that this serves as a motivation for future work.

\section*{Acknowledgements}
The authors are grateful to Alicia Curth and the 3 anonymous ICML reviewers for their useful comments \& feedback on an earlier manuscript. Nabeel Seedat is funded by the Cystic Fibrosis Trust, Fergus Imrie by the National Science Foundation (NSF), grant number 1722516, and both Zhaozhi Qian and Mihaela van der Schaar by the Office of Naval Research (ONR).

\newpage
\bibliography{refs}

\begin{thebibliography}{54}
\providecommand{\natexlab}[1]{#1}
\providecommand{\url}[1]{\texttt{#1}}
\expandafter\ifx\csname urlstyle\endcsname\relax
  \providecommand{\doi}[1]{doi: #1}\else
  \providecommand{\doi}{doi: \begingroup \urlstyle{rm}\Url}\fi

\bibitem[Alaa \& van~der Schaar(2017)Alaa and van~der Schaar]{alaa2017bayesian}
Alaa, A.~M. and van~der Schaar, M.
\newblock Bayesian inference of individualized treatment effects using
  multi-task gaussian processes.
\newblock In \emph{Proceedings of the 31st International Conference on Neural
  Information Processing Systems}, pp.\  3427--3435, 2017.

\bibitem[Alaa et~al.(2017)Alaa, Hu, and Schaar]{alaa2017learning}
Alaa, A.~M., Hu, S., and Schaar, M.
\newblock Learning from clinical judgments: Semi-markov-modulated marked hawkes
  processes for risk prognosis.
\newblock In \emph{International Conference on Machine Learning}, pp.\  60--69.
  PMLR, 2017.

\bibitem[Bao et~al.(2017)Bao, Kuang, Peissig, Page, and Willett]{bao2017hawkes}
Bao, Y., Kuang, Z., Peissig, P., Page, D., and Willett, R.
\newblock Hawkes process modeling of adverse drug reactions with longitudinal
  observational data.
\newblock In \emph{Machine learning for healthcare conference}, pp.\  177--190.
  PMLR, 2017.

\bibitem[Bartsch et~al.(2007)Bartsch, Dally, Popanda, Risch, and
  Schmezer]{bartsch2007genetic}
Bartsch, H., Dally, H., Popanda, O., Risch, A., and Schmezer, P.
\newblock Genetic risk profiles for cancer susceptibility and therapy response.
\newblock \emph{Cancer Prevention}, pp.\  19--36, 2007.

\bibitem[Bellot \& van~der Schaar(2021)Bellot and van~der
  Schaar]{bellot2021policy}
Bellot, A. and van~der Schaar, M.
\newblock Policy analysis using synthetic controls in continuous-time.
\newblock \emph{In Proceedings of 38th International Conference on Machine
  Learning (ICML 2021)}, 2021.

\bibitem[Bica et~al.(2020{\natexlab{a}})Bica, Alaa, and Van
  Der~Schaar]{bica20time}
Bica, I., Alaa, A., and Van Der~Schaar, M.
\newblock Time series deconfounder: Estimating treatment effects over time in
  the presence of hidden confounders.
\newblock In \emph{Proceedings of the 37th International Conference on Machine
  Learning}, Proceedings of Machine Learning Research. PMLR,
  2020{\natexlab{a}}.

\bibitem[Bica et~al.(2020{\natexlab{b}})Bica, Alaa, Jordon, and van~der
  Schaar]{bica2020estimating}
Bica, I., Alaa, A.~M., Jordon, J., and van~der Schaar, M.
\newblock Estimating counterfactual treatment outcomes over time through
  adversarially balanced representations.
\newblock \emph{In Proceedings of 8th International Conference on Learning
  Representations (ICLR 2020)}, 2020{\natexlab{b}}.

\bibitem[Bica et~al.(2021)Bica, Alaa, Lambert, and Van
  Der~Schaar]{bica2021real}
Bica, I., Alaa, A.~M., Lambert, C., and Van Der~Schaar, M.
\newblock From real-world patient data to individualized treatment effects
  using machine learning: current and future methods to address underlying
  challenges.
\newblock \emph{Clinical Pharmacology \& Therapeutics}, 109\penalty0
  (1):\penalty0 87--100, 2021.

\bibitem[Chen et~al.(2018)Chen, Rubanova, Bettencourt, and
  Duvenaud]{chen2018neural}
Chen, R.~T., Rubanova, Y., Bettencourt, J., and Duvenaud, D.
\newblock Neural ordinary differential equations.
\newblock In \emph{Proceedings of the 32nd International Conference on Neural
  Information Processing Systems}, pp.\  6572--6583, 2018.

\bibitem[De~Brouwer et~al.(2019)De~Brouwer, Simm, Arany, and Moreau]{de2019gru}
De~Brouwer, E., Simm, J., Arany, A., and Moreau, Y.
\newblock Gru-ode-bayes: Continuous modeling of sporadically-observed time
  series.
\newblock In \emph{33rd Conference on Neural Information Processing Systems
  (NeurIPS 2019)}, 2019.

\bibitem[Gal \& Ghahramani(2016)Gal and Ghahramani]{gal2015dropout}
Gal, Y. and Ghahramani, Z.
\newblock Dropout as a bayesian approximation: Representing model uncertainty
  in deep learning.
\newblock In \emph{international conference on machine learning}, pp.\
  1050--1059. PMLR, 2016.

\bibitem[Ganin et~al.(2016)Ganin, Ustinova, Ajakan, Germain, Larochelle,
  Laviolette, Marchand, and Lempitsky]{ganin2016}
Ganin, Y., Ustinova, E., Ajakan, H., Germain, P., Larochelle, H., Laviolette,
  F., Marchand, M., and Lempitsky, V.
\newblock Domain-adversarial training of neural networks.
\newblock \emph{The journal of machine learning research}, 17\penalty0
  (1):\penalty0 2096--2030, 2016.

\bibitem[Geng et~al.(2017)Geng, Paganetti, and Grassberger]{geng2017prediction}
Geng, C., Paganetti, H., and Grassberger, C.
\newblock Prediction of treatment response for combined chemo-and radiation
  therapy for non-small cell lung cancer patients using a bio-mathematical
  model.
\newblock \emph{Scientific reports}, 7\penalty0 (1):\penalty0 1--12, 2017.

\bibitem[Gershenwald et~al.(2017)Gershenwald, Scolyer, Hess, Sondak, Long,
  Ross, Lazar, Faries, Kirkwood, McArthur, et~al.]{tnm}
Gershenwald, J.~E., Scolyer, R.~A., Hess, K.~R., Sondak, V.~K., Long, G.~V.,
  Ross, M.~I., Lazar, A.~J., Faries, M.~B., Kirkwood, J.~M., McArthur, G.~A.,
  et~al.
\newblock Melanoma staging: evidence-based changes in the american joint
  committee on cancer eighth edition cancer staging manual.
\newblock \emph{CA: a cancer journal for clinicians}, 67\penalty0 (6):\penalty0
  472--492, 2017.

\bibitem[{GPy}(since 2012)]{gpy2014}
{GPy}.
\newblock {GPy}: A gaussian process framework in python.
\newblock \url{http://github.com/SheffieldML/GPy}, since 2012.

\bibitem[Gwak et~al.(2020)Gwak, Sim, Poli, Massaroli, Choo, and
  Choi]{gwak2020neural}
Gwak, D., Sim, G., Poli, M., Massaroli, S., Choo, J., and Choi, E.
\newblock Neural ordinary differential equations for intervention modeling.
\newblock \emph{arXiv preprint arXiv:2010.08304}, 2020.

\bibitem[Hawkes(1971)]{hawkes1971spectra}
Hawkes, A.~G.
\newblock {Spectra of some self-exciting and mutually exciting point
  processes}.
\newblock \emph{Biometrika}, 58\penalty0 (1):\penalty0 83--90, 04 1971.
\newblock \doi{10.1093/biomet/58.1.83}.
\newblock URL \url{https://doi.org/10.1093/biomet/58.1.83}.

\bibitem[Hawkes \& Oakes(1974)Hawkes and Oakes]{hawkes1974cluster}
Hawkes, A.~G. and Oakes, D.
\newblock A cluster process representation of a self-exciting process.
\newblock \emph{Journal of Applied Probability}, 11\penalty0 (3):\penalty0
  493--503, 1974.

\bibitem[Hern{\'a}n et~al.(2000)Hern{\'a}n, Brumback, and
  Robins]{hernan2000marginal}
Hern{\'a}n, M.~{\'A}., Brumback, B., and Robins, J.~M.
\newblock Marginal structural models to estimate the causal effect of
  zidovudine on the survival of hiv-positive men.
\newblock \emph{Epidemiology}, pp.\  561--570, 2000.

\bibitem[Hill(2011)]{hill2011bayesian}
Hill, J.~L.
\newblock Bayesian nonparametric modeling for causal inference.
\newblock \emph{Journal of Computational and Graphical Statistics}, 20\penalty0
  (1):\penalty0 217--240, 2011.

\bibitem[Ilg et~al.(2018)Ilg, Cicek, Galesso, Klein, Makansi, Hutter, and
  Brox]{ilg2018uncertainty}
Ilg, E., Cicek, O., Galesso, S., Klein, A., Makansi, O., Hutter, F., and Brox,
  T.
\newblock Uncertainty estimates and multi-hypotheses networks for optical flow.
\newblock In \emph{Proceedings of the European Conference on Computer Vision
  (ECCV)}, pp.\  652--667, 2018.

\bibitem[Jesson et~al.(2020)Jesson, Mindermann, Shalit, and
  Gal]{jesson2020identifying}
Jesson, A., Mindermann, S., Shalit, U., and Gal, Y.
\newblock Identifying causal-effect inference failure with uncertainty-aware
  models.
\newblock \emph{Advances in Neural Information Processing Systems}, 33, 2020.

\bibitem[Johansson et~al.(2016)Johansson, Shalit, and
  Sontag]{johansson2016learning}
Johansson, F., Shalit, U., and Sontag, D.
\newblock Learning representations for counterfactual inference.
\newblock In \emph{International conference on machine learning}, pp.\
  3020--3029. PMLR, 2016.

\bibitem[Johansson et~al.(2019)Johansson, Sontag, and
  Ranganath]{johansson2019support}
Johansson, F.~D., Sontag, D., and Ranganath, R.
\newblock Support and invertibility in domain-invariant representations.
\newblock In \emph{The 22nd International Conference on Artificial Intelligence
  and Statistics}, pp.\  527--536. PMLR, 2019.

\bibitem[Johansson et~al.(2020)Johansson, Shalit, Kallus, and
  Sontag]{johansson2020generalization}
Johansson, F.~D., Shalit, U., Kallus, N., and Sontag, D.
\newblock Generalization bounds and representation learning for estimation of
  potential outcomes and causal effects.
\newblock \emph{arXiv preprint arXiv:2001.07426}, 2020.

\bibitem[Kidger et~al.(2020)Kidger, Morrill, Foster, and
  Lyons]{kidger2020neural}
Kidger, P., Morrill, J., Foster, J., and Lyons, T.
\newblock Neural controlled differential equations for irregular time series.
\newblock \emph{34th Conference on Neural Information Processing Systems
  (NeurIPS 2020)}, 2020.

\bibitem[Lee et~al.(2016)Lee, Lim, and Ong]{lee2016hawkes}
Lee, Y., Lim, K.~W., and Ong, C.~S.
\newblock Hawkes processes with stochastic excitations.
\newblock In \emph{International Conference on Machine Learning}, pp.\  79--88.
  PMLR, 2016.

\bibitem[Li \& Fu(2017)Li and Fu]{li2017matching}
Li, S. and Fu, Y.
\newblock Matching on balanced nonlinear representations for treatment effects
  estimation.
\newblock In \emph{NIPS}, 2017.

\bibitem[Lim et~al.(2018)Lim, Alaa, and van~der Schaar]{lim2018forecasting}
Lim, B., Alaa, A.~M., and van~der Schaar, M.
\newblock Forecasting treatment responses over time using recurrent marginal
  structural networks.
\newblock \emph{NeurIPS}, 18:\penalty0 7483--7493, 2018.

\bibitem[Lok(2008)]{lok2008statistical}
Lok, J.
\newblock Statistical modeling of causal effects in continuous time.
\newblock \emph{Annals of statistics}, 36\penalty0 (3):\penalty0 1464--1507,
  2008.

\bibitem[Lyons et~al.(2007)Lyons, Caruana, and L{\'e}vy]{lyons2007differential}
Lyons, T.~J., Caruana, M., and L{\'e}vy, T.
\newblock \emph{Differential equations driven by rough paths}.
\newblock Springer, 2007.

\bibitem[Mansournia et~al.(2012)Mansournia, Danaei, Forouzanfar, Mahmoodi,
  Jamali, Mansournia, and Mohammad]{mansournia2012effect}
Mansournia, M.~A., Danaei, G., Forouzanfar, M.~H., Mahmoodi, M., Jamali, M.,
  Mansournia, N., and Mohammad, K.
\newblock Effect of physical activity on functional performance and knee pain
  in patients with osteoarthritis: analysis with marginal structural models.
\newblock \emph{Epidemiology}, pp.\  631--640, 2012.

\bibitem[Mansournia et~al.(2017)Mansournia, Etminan, Danaei, Kaufman, and
  Collins]{mansournia2017handling}
Mansournia, M.~A., Etminan, M., Danaei, G., Kaufman, J.~S., and Collins, G.
\newblock Handling time varying confounding in observational research.
\newblock \emph{bmj}, 359, 2017.

\bibitem[Morrill et~al.(2021)Morrill, Kidger, Yang, and
  Lyons]{morrill2021neural}
Morrill, J., Kidger, P., Yang, L., and Lyons, T.
\newblock Neural controlled differential equations for online prediction tasks.
\newblock \emph{arXiv preprint arXiv:2106.11028}, 2021.

\bibitem[Platt et~al.(2009)Platt, Schisterman, and Cole]{platt2009time}
Platt, R.~W., Schisterman, E.~F., and Cole, S.~R.
\newblock Time-modified confounding.
\newblock \emph{American journal of epidemiology}, 170\penalty0 (6):\penalty0
  687--694, 2009.

\bibitem[Robins(1986)]{robins1986new}
Robins, J.
\newblock A new approach to causal inference in mortality studies with a
  sustained exposure period—application to control of the healthy worker
  survivor effect.
\newblock \emph{Mathematical modelling}, 7\penalty0 (9-12):\penalty0
  1393--1512, 1986.

\bibitem[Robins et~al.(2000)Robins, Hern{\'a}n, and
  Brumback]{robins2000marginal}
Robins, J., Hern{\'a}n, M., and Brumback, B.
\newblock Marginal structural models and causal inference in epidemiology.
\newblock \emph{Epidemiology (Cambridge, Mass.)}, 11\penalty0 (5):\penalty0
  550--560, 2000.

\bibitem[Robins(1994)]{robins1994correcting}
Robins, J.~M.
\newblock Correcting for non-compliance in randomized trials using structural
  nested mean models.
\newblock \emph{Communications in Statistics-Theory and methods}, 23\penalty0
  (8):\penalty0 2379--2412, 1994.

\bibitem[Robins(1997)]{robins1997}
Robins, J.~M.
\newblock Causal inference from complex longitudinal data.
\newblock In \emph{Latent variable modeling and applications to causality},
  pp.\  69--117. Springer, 1997.

\bibitem[Rosenbaum \& Rubin(1983)Rosenbaum and Rubin]{rosenbaum1983central}
Rosenbaum, P.~R. and Rubin, D.~B.
\newblock The central role of the propensity score in observational studies for
  causal effects.
\newblock \emph{Biometrika}, 70\penalty0 (1):\penalty0 41--55, 1983.

\bibitem[Roueff et~al.(2016)Roueff, Von~Sachs, and
  Sansonnet]{roueff2016locally}
Roueff, F., Von~Sachs, R., and Sansonnet, L.
\newblock Locally stationary hawkes processes.
\newblock \emph{Stochastic Processes and their Applications}, 126\penalty0
  (6):\penalty0 1710--1743, 2016.

\bibitem[Rubanova et~al.(2019)Rubanova, Chen, and Duvenaud]{rubanova2019latent}
Rubanova, Y., Chen, R.~T., and Duvenaud, D.
\newblock Latent {ODE}s for irregularly-sampled time series.
\newblock In \emph{Proceedings of the 33rd International Conference on Neural
  Information Processing Systems}, pp.\  5320--5330, 2019.

\bibitem[Ryalen et~al.(2019)Ryalen, Stensrud, and
  R{\o}ysland]{ryalen2019additive}
Ryalen, P.~C., Stensrud, M.~J., and R{\o}ysland, K.
\newblock The additive hazard estimator is consistent for continuous-time
  marginal structural models.
\newblock \emph{Lifetime data analysis}, 25\penalty0 (4):\penalty0 611--638,
  2019.

\bibitem[Saarela \& Liu(2016)Saarela and Liu]{saarela2016flexible}
Saarela, O. and Liu, Z.
\newblock A flexible parametric approach for estimating continuous-time inverse
  probability of treatment and censoring weights.
\newblock \emph{Statistics in medicine}, 35\penalty0 (23):\penalty0 4238--4251,
  2016.

\bibitem[Schisterman et~al.(2009)Schisterman, Cole, and
  Platt]{schisterman2009overadjustment}
Schisterman, E.~F., Cole, S.~R., and Platt, R.~W.
\newblock Overadjustment bias and unnecessary adjustment in epidemiologic
  studies.
\newblock \emph{Epidemiology (Cambridge, Mass.)}, 20\penalty0 (4):\penalty0
  488, 2009.

\bibitem[Schulam \& Saria(2017)Schulam and Saria]{schulam2017reliable}
Schulam, P. and Saria, S.
\newblock Reliable decision support using counterfactual models.
\newblock \emph{Advances in Neural Information Processing Systems},
  30:\penalty0 1697--1708, 2017.

\bibitem[Shalit et~al.(2017)Shalit, Johansson, and
  Sontag]{shalit2017estimating}
Shalit, U., Johansson, F.~D., and Sontag, D.
\newblock Estimating individual treatment effect: generalization bounds and
  algorithms.
\newblock In \emph{International Conference on Machine Learning}, pp.\
  3076--3085. PMLR, 2017.

\bibitem[Tatekawa et~al.(2014)Tatekawa, Iwata, Kawaguchi, Ishikura, Baba,
  Otsuka, Miyakawa, Iwana, and Shibamoto]{volume}
Tatekawa, K., Iwata, H., Kawaguchi, T., Ishikura, S., Baba, F., Otsuka, S.,
  Miyakawa, A., Iwana, M., and Shibamoto, Y.
\newblock Changes in volume of stage i non-small-cell lung cancer during
  stereotactic body radiotherapy.
\newblock \emph{Radiation oncology}, 9\penalty0 (1):\penalty0 1--5, 2014.

\bibitem[Van~der Maaten \& Hinton(2008)Van~der Maaten and Hinton]{tsne}
Van~der Maaten, L. and Hinton, G.
\newblock Visualizing data using t-sne.
\newblock \emph{Journal of machine learning research}, 9\penalty0 (11), 2008.

\bibitem[Wager \& Athey(2018)Wager and Athey]{wager2018estimation}
Wager, S. and Athey, S.
\newblock Estimation and inference of heterogeneous treatment effects using
  random forests.
\newblock \emph{Journal of the American Statistical Association}, 113\penalty0
  (523):\penalty0 1228--1242, 2018.

\bibitem[Wang \& Blei(2019)Wang and Blei]{wang2019blessings}
Wang, Y. and Blei, D.~M.
\newblock The blessings of multiple causes.
\newblock \emph{Journal of the American Statistical Association}, 114\penalty0
  (528):\penalty0 1574--1596, 2019.

\bibitem[Yao et~al.(2018)Yao, Li, Li, Huai, Gao, and
  Zhang]{yao2018representation}
Yao, L., Li, S., Li, Y., Huai, M., Gao, J., and Zhang, A.
\newblock Representation learning for treatment effect estimation from
  observational data.
\newblock \emph{Advances in Neural Information Processing Systems}, 31, 2018.

\bibitem[Yoon et~al.(2018)Yoon, Jordon, and Van Der~Schaar]{yoon2018ganite}
Yoon, J., Jordon, J., and Van Der~Schaar, M.
\newblock Ganite: Estimation of individualized treatment effects using
  generative adversarial nets.
\newblock In \emph{International Conference on Learning Representations}, 2018.

\bibitem[Zhang et~al.(2020)Zhang, Gao, Unterman, and Arodz]{invertzhang2020}
Zhang, H., Gao, X., Unterman, J., and Arodz, T.
\newblock Approximation capabilities of neural {ODE}s and invertible residual
  networks.
\newblock In \emph{International Conference on Machine Learning}, pp.\
  11086--11095. PMLR, 2020.

\end{thebibliography}
\bibliographystyle{icml2022}

\newpage
\appendix
\onecolumn
\section{Notation Summary}

\begin{table*}[!h]
\centering
\caption{Summary of notation used in the paper}
\vspace{0.1in}
\begin{tabular}{@{}ll@{}}
\toprule
Notation       & Explanation\\ \midrule
$\textbf{X}$       & Covariate process\\ 
$A$      & Treatment process \\ 
$Y$      & Outcome path \\ 
$Y(A=a)$      & Potential outcome of $Y$\\ 
$t$      & time $t$\\ 
$N$      & Counting Process\\ 
$\mathcal F_{t}$    & Filtration \\
$\lambda$    & Intensity process \\
$\textbf{Z}$    & Latent path \\
$f_{\theta}$    &  Encoder latent vector field\\
$f_{\phi}$    & Decoder latent vector field \\
$\textbf{X}_t$       & Covariates at time $t$\\ 
$A_t$      & Treatment at time $t$\\ 
$Y_t$      & Outcome at time $t$\\ 
$h_v$      & Outcome prediction network\\ 
$h_a$      & Treatment prediction network\\ 
$\mu$      & Parameter controlling the trade-off
between treatment and outcome prediction \\ 
$\mathcal{L}^{(a)}$ & Outcome loss\\ 
$\mathcal{L}^{(a)}$  & Treatment loss\\

 \\ \bottomrule
\end{tabular}
\label{notation}
\end{table*}

Notation for the tumor growth dynamics can be found in Appendix \ref{appendixB1}

\section{Extended Related Work}\label{appendixA}

\textbf{Treatment effect estimation with static data.}\\
A large body of work has proposed causal inference methods applied to observational data in the static setting. The methods can be grouped as:
(a) Representation learning  \cite{johansson2016learning, shalit2017estimating, yoon2018ganite, li2017matching, yao2018representation}, (b) Matching and Re-weighting \cite{rosenbaum1983central} and (c) Tree-based \cite{hill2011bayesian, wager2018estimation}.

While the aforementioned methods showed good performance, the static setting has two properties that simplify the estimation, namely (1) the treatment is allocated only once and does not change over time, and (2) the confounders and the outcomes are static variables rather than time series. 
Additionally, the static setting does not naturally apply to the medical studies based on longitudinal data (e.g. EHRs), where the treatments, confounders and outcomes are all time-varying \cite{hernan2000marginal, schisterman2009overadjustment,mansournia2012effect}. The wealth of information being routinely collected as a part of the electronic health record (EHR) provides an unprecedented opportunity to discover appropriate clinical recommendations for patients given 

\textbf{Treatment effect estimation with longitudinal, time-varying data.}\\
A number of methods have been proposed for estimating the effect of time-varying exposures in the presence of time-dependent confounding, including $g$-computation \cite{robins1986new}, Structural Nested Models \cite{robins1994correcting}, and Marginal Structural Models (MSMs) \cite{robins2000marginal}.  However, MSMs are sensitive to model mis-specification, when computing the propensity weights and hence when estimating outcomes. 

Methods for the longitudinal setting with temporal confounding such as 
Marginal Structural Models (MSMs) \cite{robins2000marginal}, have thus been extended to address some of these issues.   The Recurrent Marginal Structural Model (RMSN) \cite{lim2018forecasting} aimed to reduce variance in the weights using recurrent neural networks to estimate the propensity weights and to build the outcome model.

Recently, the Counterfactual Recurrent Network (CRN) \cite{bica2020estimating} combined a RNN architecture with adversarial training to balance the covariate distributions in different treatment regimes.  That said, the aforementioned methods applied to longitudinal, time-varying data make unrealistic assumptions that the measurements are made at regular intervals, fully aligned across all patients, or do not contain missing values. 

The Counterfactual Gaussian Process (CGP) \cite{schulam2017reliable} made no such assumptions on data sampling using a generative model to jointly model the outcome and the treatment allocation. However, CGP makes strong assumptions about model structure by using Bayesian non-parametric methods, which makes them unsuitable to handle heterogeneous treatment effects arising from baseline variables or settings with multiple treatment outcomes. In addition, CGP does not directly address the issue of time-dependent confounding.

While in this work we assume no hidden confounding, we do note a body of work such as \citet{wang2019blessings} and \citet{bica20time} exists which attempts to adjust for hidden confounders in observational data.

\textbf{Neural differential equations.}\\
We consider the trajectory a patient under-going treatment as a dynamical system. Neural ordinary differential equations (Neural ODEs) and extensions have been shown to successfully model the continuous-time evolution of dynamical systems with differential equations learned by neural networks \cite{chen2018neural,rubanova2019latent,de2019gru}. However, a key limitation of Neural ODE approaches is that the entire trajectory is determined by the initial condition at $t=0$ and the equation does not account for any information available at $t>0$.

Recent works on neural controlled differential equations (Neural CDE) \cite{kidger2020neural,morrill2021neural} address this shortcoming by allowing incoming information to modulate the dynamics. This ability is naturally useful in a clinical setting, as not only can we model the continuous-time latent state evolution of a patient trajectory, but also we account for incoming data (e.g. treatment changes) that modulate the dynamics of system.

As was discussed in the main paper, \cite{gwak2020neural} and \cite{bellot2021policy} have used neural differential equations for intervention modeling. While somewhat related by virtue of using neural differential equations, these works are not applicable to our setting.

\cite{gwak2020neural} used two separate neural ODEs to model the intervention and outcome processes. That said, the method is not applicable to treatment effect estimation in clinical settings as it does not integrate time-varying covariates or adjust for confounding. 

On the other hand  \cite{bellot2021policy} proposed to model treatment effects in continuous-time in the context of synthetic controls. The setting however, is completely different as it only considered a single intervention and the synthetic control approach is not applicable more generally. 

As a consequence, to the best of our knowledge TE-CDE is the first method for counterfactual estimation and treatment effects to leverage the mathematics of neural differential equations and more specifically neural controlled differential equations .

\newpage

\section{Simulation Environment}\label{appendixB}

\subsection{Pharmacokinetic-Pharmacodynamic Model}\label{appendixB1}

Pharmacokinetic-Pharmacodynamic models are bio-mathematical models which represent dose-response relationships. This enables clinicians to understand possible patient response and hence propose optimal treatments.

In our setting we wish to estimate counterfactual outcomes. Since counterfactuals do not exist in practice, we make use of a synthetic data generated via a PK-PD model. In particular, we use a state-of-the-art bio-mathematical by \cite{geng2017prediction}. 

The PK-PD model is used to model lung cancer tumor growth under effects of chemotherapy and radiotherapy. For evaluation of counterfactuals such a model allows us to generate data of patient tumor volume outcomes under different possible treatments.

	\textbf{Model of tumor growth} \\
	The tumor volume at time $t$ after diagnosis is modeled as follows:
\begin{equation}
\frac{dV(t)}{dt}=\bigg(\underbrace{\rho \log \left(\frac{K}{V(t)}\right)}_{\text {Tumor growth }}-\underbrace{\beta_{c} C(t)}_{\text {Chemotherapy }}-\underbrace{\left(\alpha_{r} c(t)+\beta_{r} c(t)^{2}\right)}_{\text {Radiotherapy }}+\underbrace{e_{t}}_{\text {Noise }} \bigg) V(t),
\label{eq:pkpd-app}
\end{equation}

The parameters $K, \rho, \beta_c, \alpha_r, \beta_r$ for each simulated patient are sampled from the prior distributions described in \cite{geng2017prediction} and details are outlined in Table 3 below. We include $e_t \sim \mathcal{N}(0, 0.01^2)$ as random noise to account for randomness in the tumor growth.

\begin{table}[h]
    \centering
        \caption{Outline of different parameters used in the PK-PD Model}
\scalebox{0.9}{
\begin{tabular}{ccccc}
    \toprule

 Model & Variable & Parameter & Distribution &  Parameter Value ($\mu, \sigma$))  \\
\midrule
\multirow{2}{*}{Tumor growth}
     & Growth parameter    & $\rho$ & Normal & $7.00 \times 10^{-5}$, $7.23\times 10^{-3}$\\
  &  Carrying capacity     & $K$ & Constant & 30 \\

\midrule
\midrule
\multirow{2}{*}{Radiotherapy}
     & Radio cell kill ($\alpha$)    & $ \alpha_r$ & Normal & 0.0398, 0.168\\
  & Radio cell kill ($\beta$)    & $\beta_r$ & - & Set s.t. $\alpha/\beta$=10 \\

  \midrule
\midrule
\multirow{1}{*}{Chemotherapy}
     & Chemo cell kill    & $\beta_c$ & Normal & 0.028, 0.0007\\

  \bottomrule
\end{tabular}}
    \label{table:params}
\end{table}

\textbf{Heterogeneous responses.} 
In the simulation we wish to incorporate heterogeneity among patient responses to match the real-world where individualized treatment effect can vary due to factors such as gender or genetics \cite{bartsch2007genetic}.

Similar to \cite{bica2020estimating} and \cite{lim2018forecasting}, the means are adjusted for $\beta_c$ and  $\alpha_r$ by creating three groups of patients (i.e. to represent three types of patients with heterogeneity in treatment response). 

For patient group 1, we update the mean of $\alpha_r$ such that $\mu(\alpha_r)=1.1 \times \mu(\alpha_r)$ and for patient group 3, we update the mean of $\alpha_c$ such that $\mu(\alpha_c)=1.1 \times \mu(\alpha_c)$. 

\textbf{Drug concentrations ($C(t)$ and $d(t)$).} \\
The chemotherapy drug concentration $C(t)$ follows an exponential decay relationship with a half life of one day:
	 \begin{equation}
	     C(t) = \tilde{C}(t) + C(t-1)/2,
	 \end{equation}
where $ \tilde{C}(t)$ represents a $5.0 \textnormal{mg}/\textnormal{m}^3$  dosage of Vinblastine given at time $t$. 

The radiotherapy concentration $d(t)$ represents  $2.0 Gy$ fractions of radiotherapy given at timestep $t$, where Gy is the Gray ionizing radiation dose.
	
\newpage
\textbf{Time-dependent confounding.} \\
We incorporate time-dependent confounding into the data generating process by modeling chemotherapy and radiotherapy assignment as Bernoulli random variables, with probabilities $p_c$ and $p_r$ which depend on tumor diameter: 
	\begin{align}
	p_c(t) = \sigma \left(\frac{\gamma_c}{D_{\max}} (\bar{D}(t) - \delta_c )  \right) && p_r(t) = \sigma \left(\frac{\gamma_r}{D_{\max}} (\bar{D}(t) - \delta_r ) \right),
	\end{align}
	where $D_{\max}=13 \textnormal{cm}$ is the maximum tumor diameter, $\theta_{c}=\theta_{r}=D_{\max}/2$ and $\bar{D}(t)$ is the average tumor diameter. 
The degree of time-dependent confounding is controlled by $\gamma_{c}$ and $\gamma_{r}$, where increasing $\gamma_{\{c,r\}}$ increases the amount of the time-dependent confounding.

\subsection{Cancer Staging}\label{appendixB2}
As described previously, patient outcomes can evolve through different states of disease over the course of their observational trajectory. These different states correspond to clinical guidelines used to grade cancer stages. In clinical settings, patient monitoring or observation frequency often varies for different clinical stages. 

For example, patients in more severe states with greater tumor volume are monitored more frequently and hence have a greater degree of sampling. However, as the patient's state/condition improves the measurements reduce in frequency (i.e. routine and less frequent follow-ups). 

\begin{figure*}[ht]
  \centering
  \subfigure[Changing sampling as tumor diameter varies across different stages]{\includegraphics[width=0.4\textwidth]{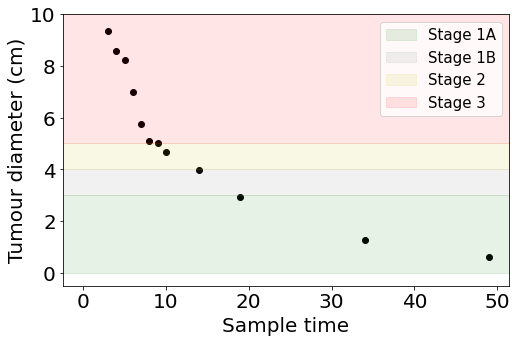}}\quad
  \subfigure[Consistent sampling for a lower cancer stage ]{\includegraphics[width=0.4\textwidth]{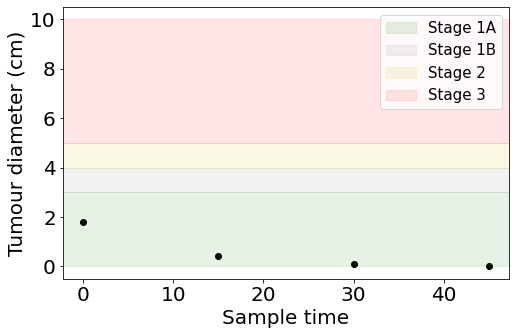}}
  \caption{Illustrations of different potential sampling patterns}
  \label{fig:sampling}
\end{figure*}

\begin{figure*}[ht]
  \centering
  \subfigure[Oscillation in tumor diameter and stage for more severe disease]{\includegraphics[width=0.4\textwidth]{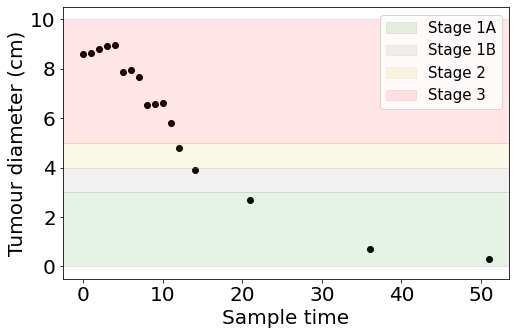}}\quad
  \subfigure[Oscillation in tumor diameter and stage for less severe disease]{\includegraphics[width=0.4\textwidth]{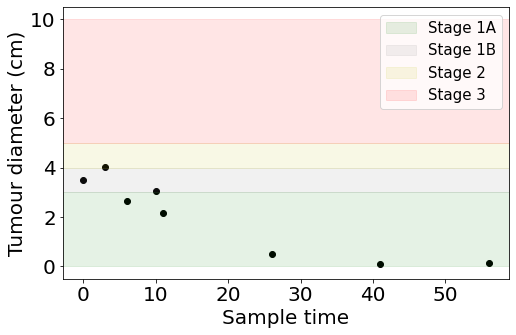}}
  \caption{Illustration that tumor diameter can oscillate and even increase over a trajectory - for example due to non-response to treatment.}
  \label{fig:change}
\end{figure*}

Thus, in the simulated data, the observational sampling rates closely follow this pattern to reflect healthcare data, shown in Figures \ref{fig:sampling} and \ref{fig:change}. This induces sampling irregularity both within a trajectory as the patient state changes over time as well as irregularity between different patient trajectories. 

This irregularity is modeled via a Hawkes process as discussed in the main paper. Note, however, that $\alpha/\beta<1$ in the Hawkes process, thereby ensuring local stationarity of the Hawkes process \cite{roueff2016locally}.

We categorize tumor stages as per the American Joint Committee on Cancer (AJCC) Tumor, Node, Metastasis (TNM) staging system \cite{tnm}: 
\begin{itemize}
    \item Stage 0: No cancer
    \item Stage 1A: 0cm $<$ tumor $\leq$ 3cm
    \item Stage 1B: 3cm $<$ tumor $\leq$ 4cm
    \item Stage 2: 4cm $<$ tumor $\leq$ 5cm
    \item Stage 3 \& 4 \footnote{We group these stages as stage 4 simply refers to the tumor invading a specific structure or anatomical space}. : tumor $>$ 5cm
\end{itemize}

The observational outcome is the volume of the lung cancer tumor, however the classification of stage is based on tumor diameter. Hence, we assume that the tumor has a perfectly spherical tumor shape similar to \cite{volume} and hence can compute the diameter as per Equation \ref{eq:volume}.

\begin{equation}
    volume = \pi/6 \times (diameter)^{3}
\label{eq:volume}
\end{equation}

\newpage
\clearpage

\section{TE-CDE}\label{appendixC}

In this section, we highlight some theoretical nuances and explanations of TE-CDE not captured in the main paper, as well as outlining further implementation details. 

\subsection{Additional Methodological Details}

\textbf{TE-CDE latent path construction}:\\
In order to construct the neural controlled differential equation a control path $X_t$ must be defined. The vector field trajectory is driven by (i.e. controlled) by the time-varying $X_t$. Recall we have an observational time series of multi-dimensional data points $((x_{t_0}, a_{t_0}, y_{t_0}), (x_{t_1}, a_{t_1}, y_{t_1}),...(x_{t_k}, a_{t_k}, y_{t_k}))$.  The control path $X_t$ is then a continuous approximation of these values, which can be obtained through interpolation (choice of interpolation is discussed later).

The latent state $z_t$ is then the solution of a CDE, controlled by $X_t$ described formally by Equation 5 and 6 below.

\begin{align}
\label{cdeeq}
     \mathbf Z_{t_0} = g_\eta(\mathbf X_{t_0}, A_{t_0}, Y_{t_0}),\\
     \mathbf Z_t =  \mathbf Z_{t_0} + \int_{t_0}^{t} f_{\theta}(\mathbf Z_s)\frac{d[\mathbf X_s,A_s,Y_s]}{ds}ds,
\end{align}
where the integral is a Riemann-Stieltjes integral.\\

\textbf{Differences between Neural CDEs and Neural ODEs}:\\
 The primary difference between a Neural CDE and a Neural ODE lies with the control path $X_t$. The control path addresses a shortcoming of a neural ODE, where the latent state $z_t$ in an ODE is decided by $z_{t0}$, hence is unsuitable to adapt to incoming data due to reliance on the initial value. Instead, neural CDEs have a latent state $z_t$ controlled by another time series $X_t$. That said, Neural CDEs can be thought of as a generalization of a neural ODE, as neural CDEs can reduce to neural ODEs for the specific case where $X_t=t$.\\
 
\textbf{Training efficiency}:\\
TE-CDE and neural CDEs in general make use of the adjoint backpropagation method. This means that the memory requirements when training TE-CDE is $\mathcal{O}(T+V)$ where T=$t_1-t_0$ (time spacing) and V is the size of the vector field.\\

\textbf{Selecting an interpolation method}:\\
Since we are performing counterfactual estimation, we thus need to retain a causal interpretation from TE-CDE. When constructing the continuous control path $X_t$, we need to be careful as all interpolation methods except Linear and Rectilinear are non-causal in nature. i.e. other interpolation methods need future points in time to approximate the signal. This means that when using Neural CDEs in scenarios that require causal interpretations, such as TE-CDE, linear or rectilinear interpolation must be used.\\

\textbf{TE-CDE solves a well-posed problem}:\\
We highlight that TE-CDE solves a well-posed problem. It was shown by \cite{lyons2007differential} that CDEs solve a well-posed problem under the condition of Lipschitz continuity. We aim to show that TE-CDE still fits the well-posed paradigm. In our architecture we make use of ReLU and softmax activations as well as common neural network layers such dropout which are Lipschitz constant. Thus, the Lipschitz conditions still hold which means that TE-CDE fulfills the following conditions of a well-posed problem: (1) a solution exists, (2) the solution is unique, (3) the behaviour of the solution changes continuously based on the input data. By virtue of TE-CDE solving a well-posed problem, we note that the training of TE-CDE is thus theoretically stable.  

\textbf{Representing data continuously}:\\
TE-CDE requires a continuous-time embedding of the observational data. Whilst this has numerous benefits, it also introduces challenges. The benefit is firstly it allows us to naturally handle irregularly sampled data 
and secondly, TE-CDE produces a solution (i.e. the latent state $z_t$) defined at all points in time (i.e. continuously). 

The challenge, however, is that there may be discontinuities in $A_{\le t}$ (e.g. treatment is applied in discrete stages). This results in a piecewise smooth control path (not smooth due to linear interpolation). Thus, in our case when using an adaptive solver with a piecewise smooth control path, we then need to explicitly inform the solver about the jumps between pieces so that its integration steps may align with them. This can be achieved using the \texttt{jump\_t} argument in the CDE solver \cite{chen2018neural,kidger2020neural}. 

If this is not done the 
adjoint backpropagation method with an adaptive ODE solver can be numerically and computationally expensive (i.e. slower), as the solver must then locate the discontinuities and then slow down to resolve them.

\subsection{TE-CDE Implementation Details}
We discuss the implementation details for TE-CDE in terms of encoder-decoder architecture, optimization and general comments about training.

\textbf{Encoder-Decoder Architecture:} \\The encoder and decoder of TE-CDE both use Neural CDEs with the same architectures - the difference being the observations for the encoder and decoder (i.e. different inputs). The integrand of the neural CDE (i.e. $f$) is a 2-layer neural network with hidden states of size=128. The dimensionality of the latent state $z$ is 8. We use linear interpolation when defining the control path $X_t$.

\textbf{Optimization:} \\We make use of Stochastic Gradient Descent (SGD) as implemented in the Pytorch framework. The ODE solver used to extrapolate the latent state $z_t$ is an adaptive step size solver namely the Dormand–Prince (\texttt{dopri5}) method, which is a member of the Runga-Kutta family of ODE Solvers. We make use of the \texttt{torchcde} package with \texttt{torchdiffeq} backend. As discussed to account for the discontinuities and piecewise smooth control path we make use of the \texttt{jump\_t} argument.

\textbf{General comments:} \\Both encoder and decoder are trained for 100 epochs each. That said we also include early stopping in the training protocol based on the validation loss, with patience=5. When MC Dropout is included, we use a dropout probability=0.1. The model was implemented with Pytorch and TorchCDE and was trained and evaluated on a single Nvidia P100 or T4 GPU.  We did not explicitly tune hyperparameters (activation function, ODE Solver) for performance. Further tuning could be done for these aspects based on a validation set. However, we did tune the learning rate (lr = 1e-3, \textbf{1e-4}$^{*}$, 1e-5, 1e-6) based on performance on the validation set.

\newpage
\section{Implementation details: Benchmark algorithms}\label{appendixD}

Due to the underlying usage of Recurrent Neural Networks (RNNs) both CRN and RMSN rely on the assumption of regularly sampled data. Thus, for the irregularly sampled data setting, which effectively means un-observed data samples for different discrete time steps,  as is commonly done we include a pre-processing step for RNN-based models, where we divide the timeline equally, interpolate and impute the ``un-observed'' observations. 

\subsection{Counterfactual Recurrent Network (CRN)}
We benchmark CRN using the Tensorflow implementation of \cite{bica2020estimating} as per \footnote{ \url{https://github.com/ioanabica/Counterfactual-Recurrent-Network}}. We pre-process the data for the irregularly sampled setting as described.

\subsection{Recurrent Marginal Structural Network (RMSN)}
We benchmark RMSN using the Tensorflow implementation of \cite{lim2018forecasting} as per \footnote{ \url{https://github.com/sjblim/rmsn_nips_2018}}. We pre-process the data for the irregularly sampled setting as described. We tune parameters as described by the authors in their repo.

\subsection{Gaussian Process (GP)}
The Gaussian Process (GP) model is evaluated as a continuous-time comparison and is inspired by \cite{schulam2017reliable}. The method however, like prior works using GPs including \cite{schulam2017reliable} does not account for time-dependent confounding. Similar to \cite{schulam2017reliable}, we use a  Matérn 3/2 kernel with variance=0.22 and lengthscale=8.0. The benchmark algorithm is implemented using GPy \cite{gpy2014}.

\newpage
\section{Additional Experiments}\label{appendixE}

We perform a number of additional experiments to further validate TE-CDE.

\subsection{Uncertainty Based Exclusion of Estimates}\label{appendix-uncertain}

Uncertainty concerning the counterfactual estimates is an important problem to consider as estimation errors can lead to harmful decisions being taken based on the incorrect estimates. Whilst \cite{jesson2020identifying} have examined uncertainty in the static setting, this area is under-studied in longitudinal setting. We aim to illustrate how uncertainty can be modeled within the TE-CDE framework. We then  evaluate the utility of the uncertainty estimates to (a) improve model outcomes and counterfactual estimates and (b) examine how uncertainty could be utilized in clinical workflows such that uncertain cases can be flagged and deferred to a clinician. 

Bayesian methods such as variational inference or MCMC can be computationally intensive. Hence, we propose to augment and re-train our TE-CDE model with MC-Dropout \cite{gal2015dropout}. This then allows us to sample $N$ counterfactual outputs $y^{(i)}$, where each dropout mask corresponds to a sample from the approximate posterior distribution $q(\theta)$. The variance of these samples is then representative of the epistemic uncertainty (model uncertainty) in counterfactual estimation. 
\begin{align}
\tilde{\textbf{y}}=\frac{1}{N}\sum y^{(i)} && \sigma^{2}=\text{Epistemic uncertainty}=var(\textbf{y})
\end{align}

We highlight the utility of the uncertainty estimates of TE-CDE to improve model outcomes and counterfactual estimates, by deferring uncertain samples. As discussed, the uncertainty is quantified by the variance of the MC-Sampled counterfactual estimates (epistemic uncertainty). The uncertainty over the trajectory at every timestep is then averaged with the uncertainty estimates represented as $u^{all} = (u_{1}^{a}, u_{2}^{a}, u_{3}^{a}...u_{n}^{a} )$. The $u^{all}$ across all patients is then rank sorted based on the uncertainty metric.

As an illustrative example, we evaluate the quality of TE-CDE's uncertainty estimates by modeling how the RMSE varies as a function of the proportion of ``uncertain'' data excluded. We posit that a useful measure of uncertainty should lead to a lower RMSE on the retained subset with highest confidence. Hence, we define a meaningful representation of uncertainty as one where the error curve decreases for increasing exclusion of uncertain data. Such a measure could then be used to improve clinical workflows by deferring uncertain estimates to clinicians.

The aforementioned test ascertains if the uncertainty estimate can correctly order the counterfactual estimates. However, a limitation as a result of MC Dropout is the coupling with the model which introduces an implicit dependency on the model that generates the prediction's and uncertainty estimates. Hence, we cannot directly compare different approaches.

To address the coupling we utilize the concept of Sparsification from \cite{ilg2018uncertainty}. This is a novel application from optical flow to causal inference. We define an Oracle ($u^{oracle} = (u_{1}^{o}, u_{2}^{o}, u_{3}^{o}...u_{n}^{o} )$). The oracle has the best possible ordering of samples, with bounds based on knowing the ground truth and thereby ordering by magnitude of the true error. Thus, the oracle is the lower bound on performance. We can then compute the difference between uncertainty based exclusion and the oracle exclusion given by Equation 8. The oracle is model dependant and by computing the difference we marginalize out the impact of the oracle.  Hence, we can then use the sparsification error to directly compare different uncertainty estimates.

\begin{equation}
    \text{Sparsification error} =\sum_{i=1}^{n} u_{i}^{a}-u_{i}^{o}
\end{equation}

In order, to produce a single comparative metric to compare models, the area under the sparsification error curve (AUSE) can be used to compare the different models and uncertainty measures. Smaller values correspond to  better performance as the measure would then be closer to the oracle lower bound (i.e. closer to ground truth).

\textbf{Baseline:} We use two baselines to validate the utility of TE-CDE's uncertainty estimates: (1) Oracle: best possible exclusion and (2) Random: randomly exclude a proportion of samples. Random exclusion should not result in a reduction in RMSE.

\begin{figure*}[h]
  \centering
  \subfigure[Exclusion based on uncertainty: decreased error as more uncertain samples are excluded. The result highlights that a small proportion of samples are responsible for most of the error]{\includegraphics[width=0.4\textwidth]{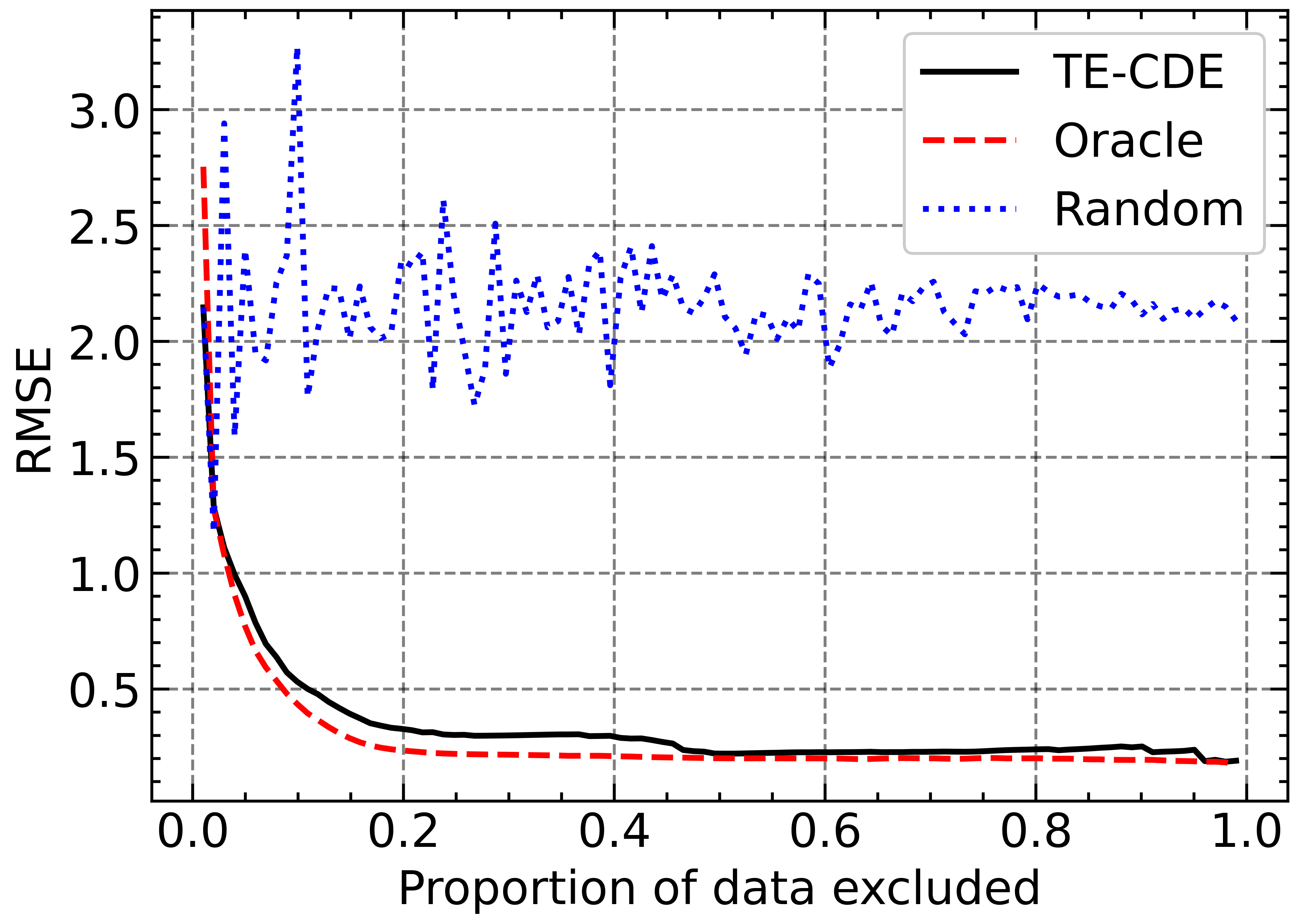}}\quad
  \subfigure[Sparsification curve with AUSE]{\includegraphics[width=0.4\textwidth]{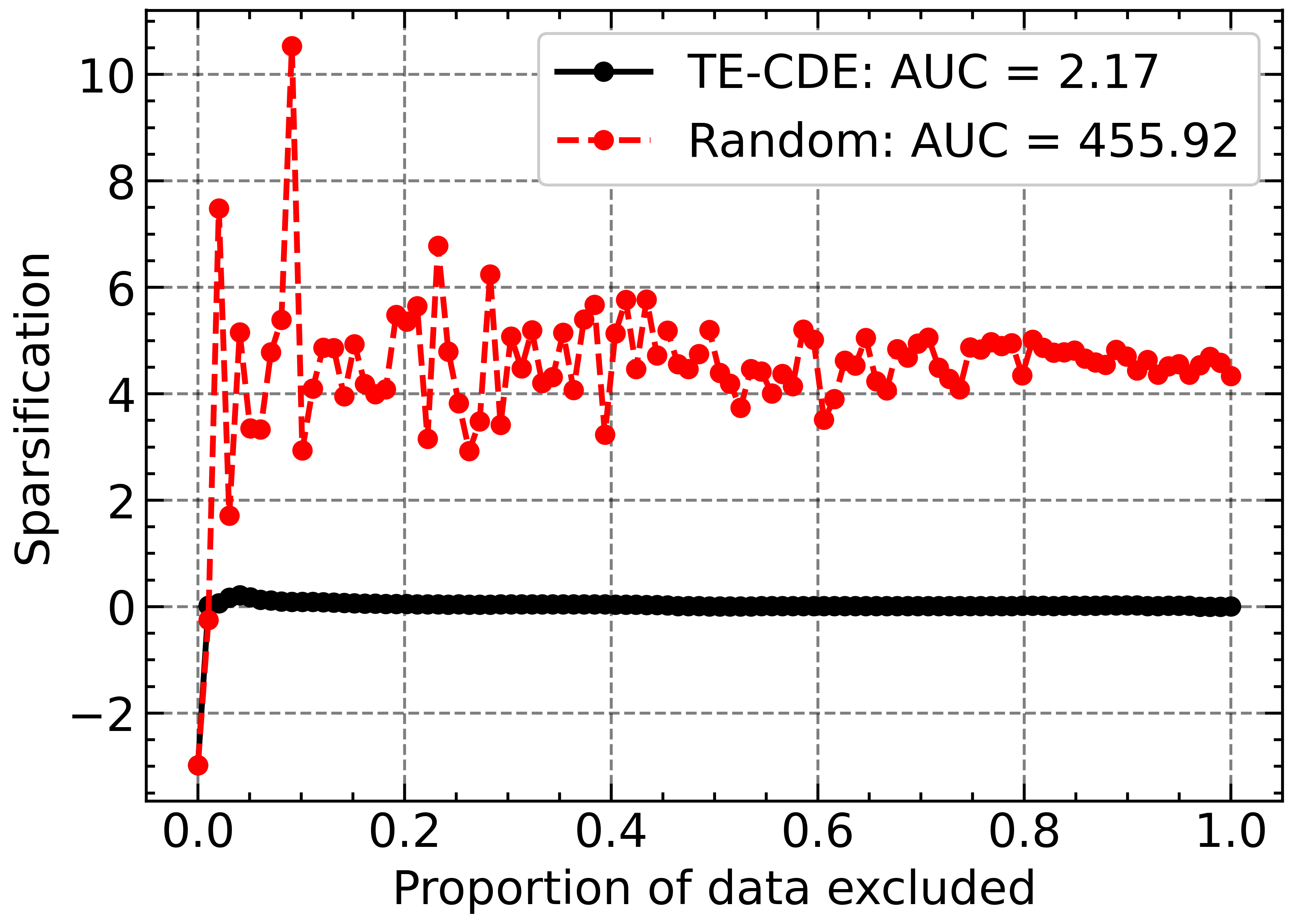}}
  \caption{Evaluation of the uncertainty representations for TE-CDE based on exclusions and sparsification}
  \label{fig:uncertainty}
\end{figure*}

\textbf{Results:} The exclusion curves are illustrated in Figure \ref{fig:uncertainty} (a) demonstrates a decreasing curve for sorting by variance/epistemic uncertainty. This suggests TE-CDE can be used to produce a useful ranking by uncertainty and performs better than the random baseline. This is corroborated by the Sparsification curve and AUSE (Figure \ref{fig:uncertainty} (b)). 

When comparing exclusion methods it highlights that ordering by the variance/epistemic uncertainty significantly outperforms random exclusion. Moreover, performance is close to the oracle, thereby suggesting TE-CDE can be used to obtain a useful measure of uncertainty. This can then be used to exclude uncertain samples and improve performance. An important observation from a clinical perspective is that a small percentage ($<10\%$) of samples are responsible for the majority of the overall estimation error. This highlights the utility that by deferring uncertain samples to clinicians we could significantly improve overall predictive and patient outcomes.

\subsection{Data Efficiency Benchmark for Different Methods}\label{appendix-efficiency}
Data efficiency is critical in clinical scenarios, where we may not have large datasets. Hence, we wish to quantify the sensitivity of the different methods to the size of the training set.

We contrast the RMSE for counterfactual estimation of TE-CDE, CRN and RMSN for varying amounts of training data,  $n=1000,5000$ relative to $n=10000$. We fix the level of time-varying confounding as $\gamma$ = 2 and $\kappa=10$. 

The results in Table \ref{table:efficiency} highlight that TE-CDE achieves the lowest RMSE percentage drop-off for all amounts of training data. Whilst, all methods demonstrate reduced performance as the amount of training data decreases, the performance degradation is less severe for TE-CDE compared to CRN and more so when compared to RMSN. This demonstrates that TE-CDE is more data efficient, which is valuable in clinical settings where the data volume may be limited.  

We posit the reason for this phenomenon is that since TE-CDE operates in continuous-time and has a $z_{t}$ defined over all time points, that less data is required to learn the underlying dynamics. Whereas, both CRN and RMSN in the discrete paradigm would need more ``discrete'' examples in order to learn the underlying dynamics for different time steps.

\begin{table}[!h]
    \centering
        \caption{Data efficiency of models trained on varying amounts of training samples for fixed $\gamma=2$}
\scalebox{0.9}{
\begin{tabular}{ccc}
    \toprule

  &  & \% Performance reduction (RMSE) vs. \\
 Train samples & Model & 10000 train samples (lower better)  \\
\midrule
\multirow{3}{*}{5000}
     & \textbf{TE-CDE}    & \textbf{4.8\%} \\
  &  CRN     & 6.4\% \\
  &  RMSN     &  25.6\% \\

\midrule
\midrule
\multirow{3}{*}{1000}
    & \textbf{TE-CDE}   & \textbf{17.1\%} \\
  &  CRN     & 18.9\% \\
  &  RMSN    & 57.2\% \\

  \bottomrule
\end{tabular}}
    \label{table:efficiency}
\end{table}

\newpage

\subsection{Latent Representation and Discovery}\label{appendix-latent}

\subsubsection{Treatment Invariant Latent Representation}
\begin{figure*}[h]
  \centering
 \subfigure[t=10]{\includegraphics[width=0.3\textwidth]{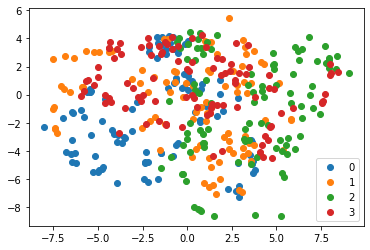}}
  \subfigure[t=30]{\includegraphics[width=0.3\textwidth]{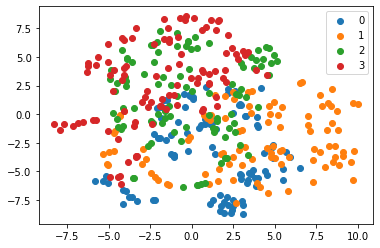}}\quad
  \subfigure[t=50]{\includegraphics[width=0.3\textwidth]{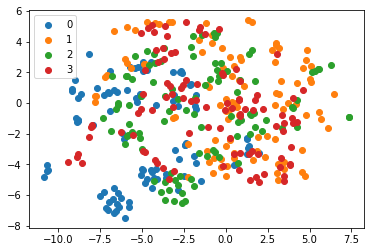}}
  \caption{T-SNE low-dimension embedding of the latent representation of TE-CDE, i.e $z_{t}$ at different timesteps. It highlights that the representation is invariant of treatment, hence assists to deal with time-dependent confounding originating from treatment history $H_{t}$. The different color dots represent different treatment options}
  \label{fig:latent_treatment}
\end{figure*}

We wish to evaluate whether the latent representation $z_{t}$ over time has indeed learnt treatment invariant representations. 
Figure \ref{fig:latent_treatment} illustrates via low dimensional T-SNE embeddings that the TE-CDE latent state learns treatment invariant representations for different treatments over different time periods i.e. with no treatment, chemotherapy, radiotherapy, combined chemotherapy and radiotherapy.

\subsubsection{Discovery and Insights from Latent Representations}

As demonstrated in Appendix \ref{appendix-uncertain}, we can ascertain which samples we are most uncertain about and use that to exclude/defer those samples. Additionally, it was highlighted that a small proportion of samples are responsible for the overall error.

From a clinical discovery perspective it is useful to ascertain unique characteristics of the uncertain samples and from a representation perspective whether there is a difference between the certain and uncertain samples. For the experiment we obtain the most certain and uncertain samples as characterized from the test set. Additionally, we obtain the most uncertain samples, which contributed most to the error from Figure 8(a). Understanding the nature of these samples is clinically beneficial.

\begin{figure}[h]
\centering
    \includegraphics[width=0.37\textwidth]{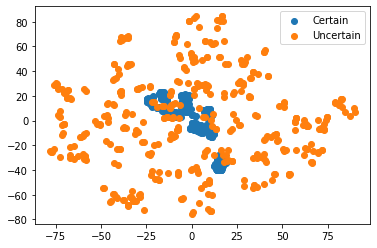}
  \caption{T-SNE of the latent representation for both certain (blue) and uncertain (orange) samples}
  \label{fig:latent}
\end{figure}

We perform a T-SNE embedding \cite{tsne} of the latent representation ($z$) for the certain and uncertain samples at the final time-step. As illustrated in Figure \ref{fig:latent}, there is a marked difference between the certain and uncertain samples. Additionally, when comparing the mean tumor volumes for each group (i.e. indicative of severity), it is evident that the uncertain group consisted of more severe patients when compared to the certain group. Furthermore, the uncertain group had both greater mean and greater variability, where Uncertain volume: 5.37 $\pm$ 20 cm$^{3}$ and Certain volume: 2.57 $\pm$ 11 cm$^{3}$. These results highlight the distinct difference between the certain and uncertain samples. This demonstrates that not only do the latent representations capture this difference and that the uncertainty estimates are useful, but also that we can glean clinically useful information from a discovery perspective from the latent representation of TE-CDE.

\newpage

\subsection{Linear $f$ and $g$}
We conduct an ablation where $f$ and $g$ are linear instead of neural networks. We then assess performance for different amounts of time-dependent confounding ($\gamma$ = 2, 4, 6, 8, 10). In terms of performance,  we note a drop in RMSE of approximately 0.093 ± 0.056.

This of course highlights the importance of the neural network to performance. However, the model still retains performant properties suggesting the main source of performance gain is indeed from the CDE component. 

\subsection{Other assessment}

We perform an added experiment based on the MIMIC-III ICU dataset \cite{johansson2016learning}. Please note the caveat we list below. Similar to \citet{bica2020estimating}, we predict white blood cell count in response to antibiotic treatment.  Consequently, we follow the related work and test on factuals. The results show similar performance for TE-CDE and CRN (difference <0.01 normalized RMSE), with both models predicting factuals with high accuracy. Note such experiments shouldn't be used to compare models for counterfactual prediction. As for the task on real data, we do not observe counterfactuals, rather only factuals. Hence, as discussed in Section 1, this is more a time-series evaluation rather than a counterfactual outcomes evaluation; which is the focus of this paper.

\subsection{Impact of Time-Dependent Confounding Across Varying Sampling Intensities}

Table \ref{table:kappa} shows detailed results for the counterfactual predictions under varying sampling intensity ($\kappa$), across various levels of time-dependent confounding ($\gamma$). We compare TE-CDE to CRN, RMSN and also highlight results for the GP.

\begin{table}[h]
    \centering
        \caption{Detailed results for counterfactual estimation under time-dependent confounding for different $\kappa$}
\scalebox{0.8}{
\begin{tabular}{ccccccc}
    \toprule
Sampling intensity $\kappa$ & Model & $\gamma=2$ & $\gamma=4$ & $\gamma=6$ & $\gamma=8$ & $\gamma=10$  \\
\midrule
\multirow{4}{*}{$\kappa=1$}
     & \textbf{TE-CDE}   & 0.96 $\pm$ 0.01 & 1.55 $\pm$ 0.02  & 1.83 $\pm$ 0.08 & 2.26 $\pm$ 0.06 & 3.07 $\pm$ 0.35  \\
  &  CRN     & 1.07 $\pm$ 0.09 & 1.63 $\pm$ 0.10  & 2.11 $\pm$ 0.29 & 3.23 $\pm$ 0.36 & 4.49 $\pm$ 0.38 \\
  &  RMSN     & 1.92 $\pm$ 0.07 & 2.42 $\pm$ 0.11 & 2.66 $\pm$ 0.15 & 4.00 $\pm$ 0.12 & 4.16 $\pm$ 0.11 \\
  &  TE-CDE ($\mu=0$)   & 1.04 $\pm$ 0.16 & 1.77 $\pm$ 0.11 & 2.14 $\pm$ 0.34 & 3.00 $\pm$ 0.21 & 4.26 $\pm$ 0.51 \\
  &  GP  & 1.74 $\pm$ 0.76 & 2.80 $\pm$ 0.93 & 3.74 $\pm$ 0.40 & 6.15 $\pm$ 1.14 & 6.97 $\pm$ 1.02 \\

\midrule
\midrule

\multirow{4}{*}{$\kappa=5$}
     & \textbf{TE-CDE}    & 0.88 $\pm$ 0.01 &  1.31 $\pm$ 0.02  & 2.32 $\pm$  0.08 & 2.62 $\pm$  0.06 & 3.02 $\pm$ 0.35 \\
  &  CRN     & 1.27 $\pm$ 0.15 & 1.78 $\pm$ 0.14  & 3.08 $\pm$ 0.21 & 4.11 $\pm$ 0.38 & 4.90 $\pm$ 0.31 \\
  &  RMSN    & 2.69 $\pm$ 0.12 & 2.92 $\pm$ 0.24 & 3.16 $\pm$ 0.18 & 4.265 $\pm$ 0.13 & 5.40 $\pm$ 0.16 \\
    &  TE-CDE ($\mu=0$)    & 0.88 $\pm$ 0.07 & 1.68 $\pm$ 0.05  & 2.97 $\pm$ 0.20 & 4.00 $\pm$ 0.22 & 4.64 $\pm$ 0.59 \\
    &  GP  & 9.08 $\pm$ 1.53 & 14.19 $\pm$ 1.73 & 24.47 $\pm$ 2.89 & 39.89 $\pm$ 1.02 & 42.99 $\pm$ 2.76 \\

\midrule
\midrule

\multirow{4}{*}{$\kappa=10$}
     & \textbf{TE-CDE}    & 0.78 $\pm$ 0.01 & 1.16 $\pm$ 0.02 & 1.93 $\pm$ 0.08 & 2.67 $\pm$ 0.06 & 3.03 $\pm$ 0.35\\
  &  CRN     & 1.31 $\pm$ 0.10 & 1.51 $\pm$ 0.11 & 3.10 $\pm$ 0.26 & 4.28 $\pm$ 0.29 & 4.84 $\pm$ 0.31 \\
  &  RMSN     & 1.60 $\pm$ 0.11 & 2.04 $\pm$ 0.12 &  3.67 $\pm$ 0.22 & 4.62  $\pm$ 0.27 & 5.41  $\pm$ 0.31 \\
    &  TE-CDE ($\mu=0$)  & 0.85 $\pm$ 0.06 &  1.23 $\pm$ 0.07  & 2.48 $\pm$ 0.06 & 3.49 $\pm$ 0.51 & 4.94 $\pm$ 0.59 \\
&  GP  &  9.41 $\pm$ 0.98 & 14.43 $\pm$ 0.95 & 24.74 $\pm$ 1.97 & 40.21 $\pm$ 0.93 & 43.78 $\pm$ 3.11
 \\
  \bottomrule
\end{tabular}}
    \label{table:kappa}
\end{table}

\subsection{Treatment-Conditioned Sampling}
Table \ref{table:results-treatment-detailed} shows detailed results for counterfactual estimation for treatment conditioned sampling (including results for the GP). Sampling intensity $\kappa$ is conditioned on treatment such that $\kappa_{treated}=10$ and $\kappa_{untreated}=1$ and $\gamma=4$.

\begin{table}[h]
    \centering
        \caption{Comparison of different methods TE-CDE, CRN, RMSN, and GP for counterfactual estimation with $\kappa$ conditioned on treatment, where $\kappa_{treated}=10$ and $\kappa_{untreated}=1$.}
\scalebox{0.75}{
\begin{tabular}{cccc}
    \toprule

Model   &  Overall         & Treated         & Untreated       \\ \midrule \midrule
TE-CDE  & 1.18 $\pm$ 0.05  & 1.56 $\pm$ 0.06 & 0.22 $\pm$ 0.02 \\ \midrule \midrule
CRN     & 1.57 $\pm$ 0.06  & 1.97 $\pm$ 0.05 & 0.64 $\pm$ 0.08 \\ \midrule \midrule
RMSN    &  3.06 $\pm$ 0.09 & 3.12 $\pm$ 0.07 & 2.83 $\pm$ 0.08 \\ \midrule
\midrule
GP    & 11.05  $\pm$ 0.23 & 16.05 $\pm$ 0.54 & 5.66 $\pm$ 0.45 \\ \midrule
  \bottomrule
\end{tabular}}
    \label{table:results-treatment-detailed}
\end{table}

\subsection{Forecasting at Additional Time Horizons}\label{appendix-horizons}
Figure 11 (a)-(c) below shows further results for forecasting at additional time horizons. The results demonstrate the same relative performance of different methods as was shown in the main paper. 
\begin{figure*}[h]
  \centering
 \subfigure[Horizon = $t_{k+3}$]{\includegraphics[width=0.25\textwidth]{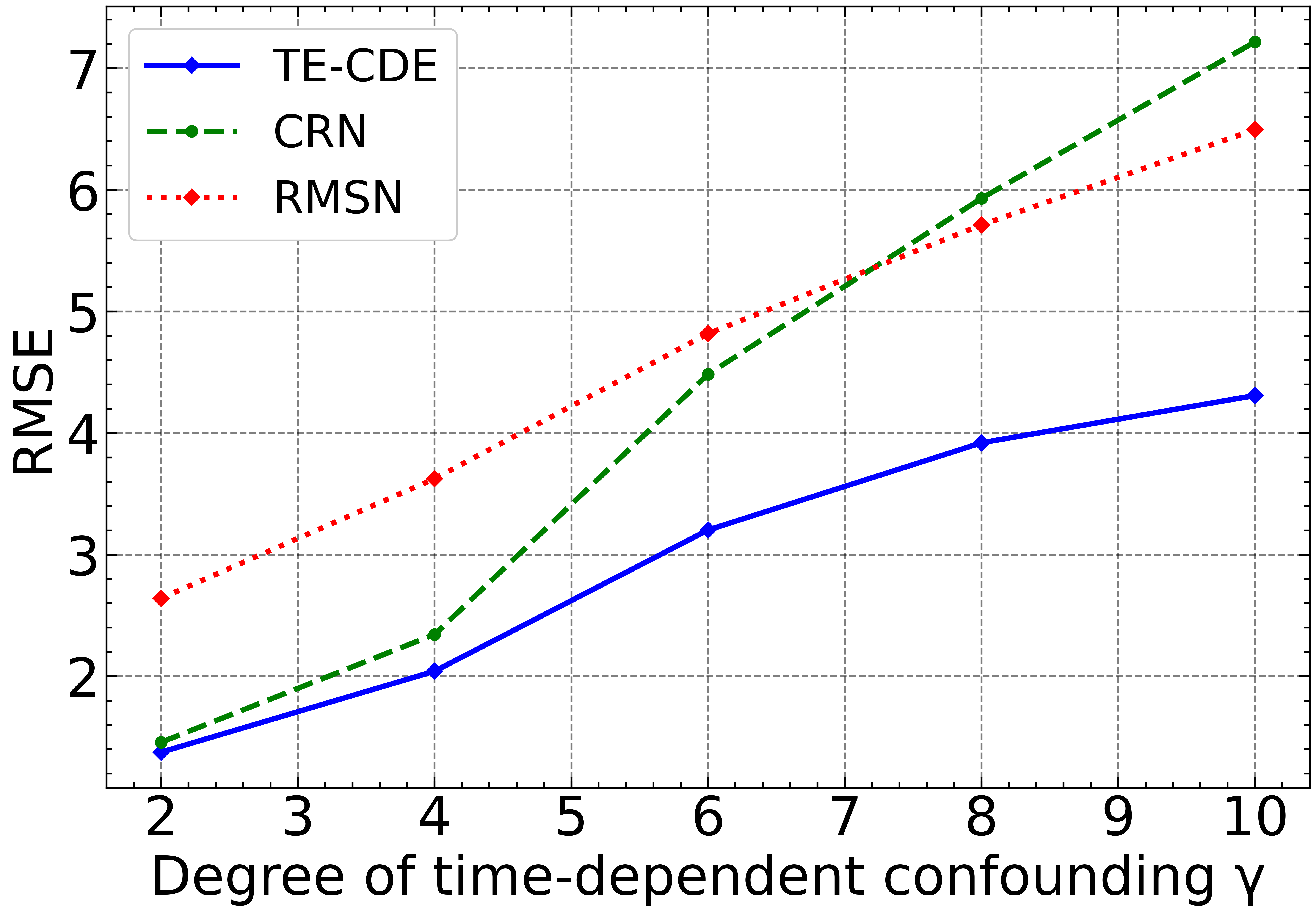}}
  \subfigure[Horizon = $t_{k+4}$]{\includegraphics[width=0.25\textwidth]{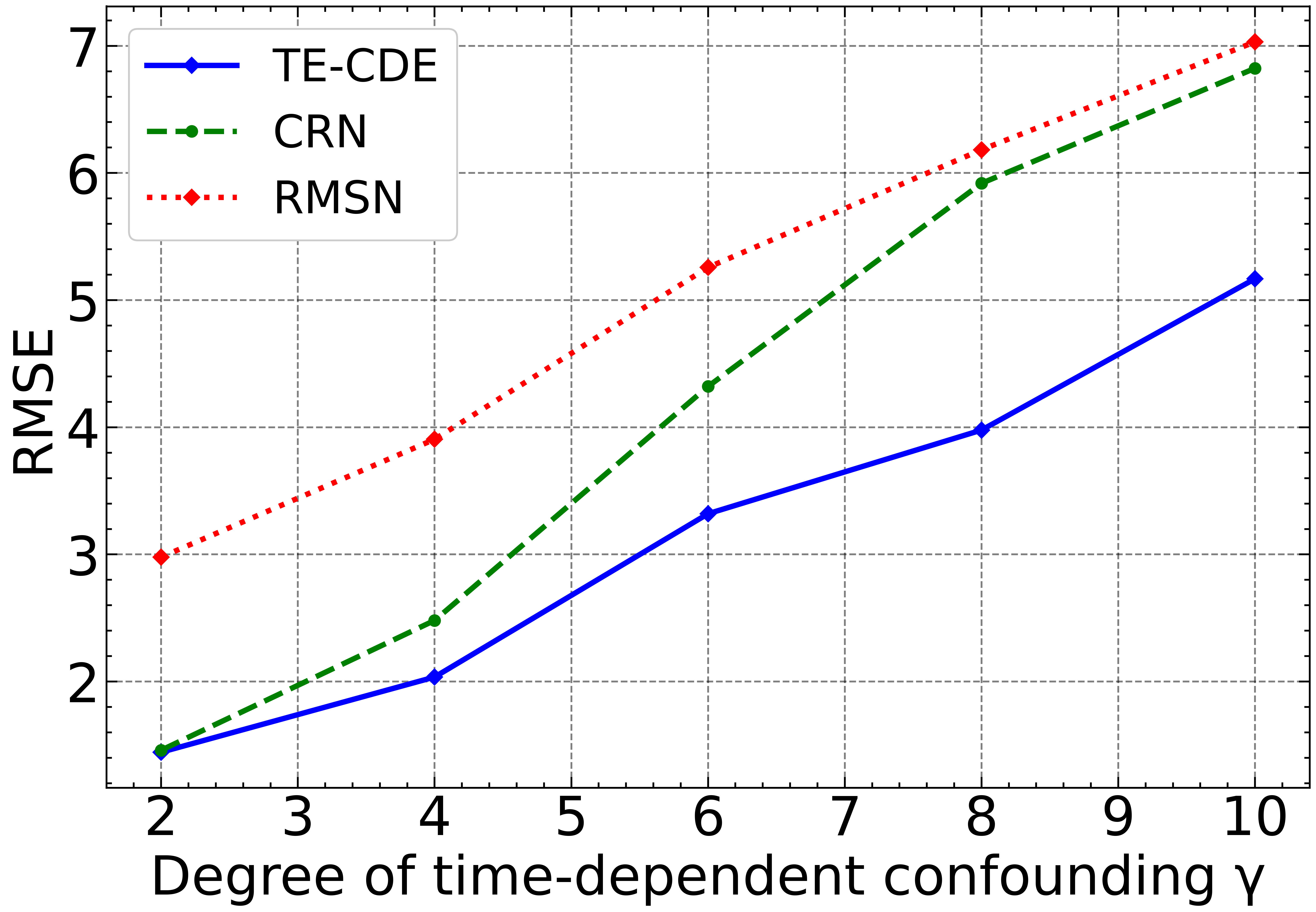}}\quad
  \subfigure[Horizon = $t_{k+5}$]{\includegraphics[width=0.25\textwidth]{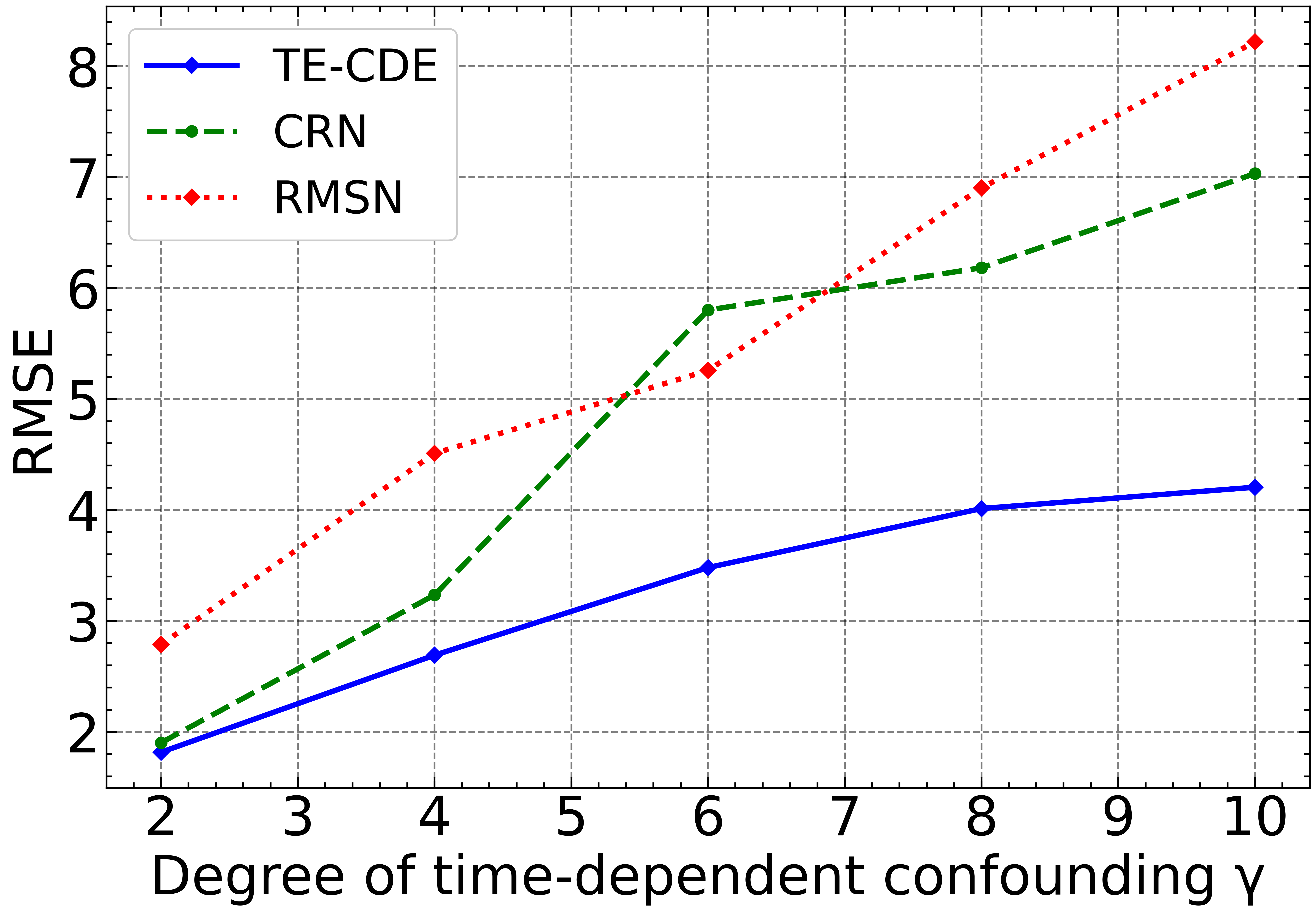}}
  \caption{Results for counterfactual estimation at additional forecasting time horizons}
  \label{fig:horizon}
\end{figure*}



\end{document}